\documentclass[11pt]{article}
\pdfoutput=1
\usepackage[margin=1in]{geometry}
\usepackage[T1]{fontenc}
\usepackage[utf8]{inputenc}
\usepackage{lmodern}
\usepackage{microtype}
\usepackage{float}
\usepackage{amsmath,amssymb,mathtools,bm}
\usepackage{graphicx}
\usepackage{booktabs,tabularx,longtable}
\usepackage{caption,subcaption}
\usepackage{xcolor}
\usepackage[most]{tcolorbox}
\usepackage{longtable}
\usepackage{tikz}
\usepackage{algorithm}
\usepackage{algpseudocode}
\usetikzlibrary{arrows.meta,positioning,decorations.pathmorphing,shapes,matrix}
\usepackage[numbers,sort&compress]{natbib}
\usepackage{hyperref}
\hypersetup{colorlinks=true, linkcolor=blue, citecolor=blue, urlcolor=blue}
\usepackage{authblk}
\usepackage{float}  

\title{Making Evidence Actionable in Adaptive Learning}
\author[1,*]{Amirreza Mehrabi}
\author[1]{Jason Wade Morphew}
\author[1]{Breejha Quezada}
\author[2]{N. Sanjay Rebello}
\affil[1]{School of Engineering Education, Purdue University, West Lafayette, IN, 47907, USA}
\affil[2]{Department of Physics and Astronomy, Purdue University, West Lafayette, IN, 47907, USA}
\affil[*]{e-mail: amehrabi@purdue.edu}
\date{}
\begin{document}
\maketitle

\flushbottom
\begin{abstract}
Adaptive learning often diagnoses precisely yet intervenes weakly, yielding help that is mistimed or misaligned. This study presents evidence supporting an instructor-governed feedback loop that converts concept-level assessment evidence into vetted micro-interventions. This adaptive learning algorithm contains three safeguards: adequacy as a hard guarantee of gap closure, attention as a priced budget for time and redundancy, and diversity as protection against overfitting to a single resource. We formalize intervention assignment as a binary integer program with constraints for coverage, time, difficulty windows from ability estimates, prerequisites encoded by a concept matrix, and anti-redundancy with diversity. Greedy selection serves low-richness and tight-latency regimes, gradient-based relaxation serves rich repositories, and a hybrid switches along a richness–latency frontier. In simulation and in an introductory physics deployment with one thousand two hundred four students, both solvers achieved full skill coverage for essentially all learners within bounded watch time. The gradient-based method reduced redundant coverage by roughly twelve percentage points relative to greedy and harmonized difficulty across slates, while greedy delivered comparable adequacy with lower computational cost in scarce settings. Slack variables localized missing content and drove targeted curation, sustaining sufficiency across subgroups. The result is a tractable and auditable controller that closes the diagnostic–pedagogical loop and delivers equitable, load-aware personalization at a classroom scale.
\end{abstract}

\section{Introduction}
The rise of adaptive learning technologies has promised a revolution where instruction bends to individual learners' needs in real time \citep{Holstein2021DesigningFair, Rodrigues2023Adaptive}. Powered by psychometric models and machine learning, these systems diagnose knowledge states, flag conceptual gaps, and recommend personalized tasks, yet their progress has been one-sided because they excel at identifying what learners lack while often overlooking how to transform diagnosis into pedagogically meaningful instruction \citep{Aleven2016Example, vanlehn2011relative}. This shortfall reflects a broader misalignment between measurement-centric architectures and the cognitive realities of learning. A misconception is evident when a learner's incorrect answer originates from a fundamental misunderstanding of a concept, rather than from non-conceptual sources such as calculation mistakes or lapses in memory. When a misunderstanding happens, most of the adaptive systems prioritize fixes the answer without repairing the underlying concept, yielding short-term gains without considering the concept understanding
\citep{Koedinger2013NewPotentials, Xie2019TrendsCompEdu}. The result is precision in error detection coupled with imprecision in instructional response, an asymmetry that raises scores while leaving misconceptions intact. At the root is a mistaken metaphor that frames learning as a search problem in which the system seeks the next item that maximizes information about hidden traits \citep{de2011generalized, Rupp2008}. 
This paradigm sharpens measurement but leaves instructional selection underspecified, producing systems that map deficits with statistical accuracy yet cannot guarantee cognitively appropriate or culturally inclusive remediation \citep{Mavrikis2015, mehrabi2024ai}. The arrival of large language models amplifies this tendency because, although such models can infer rich error patterns at scale, scale is not pedagogy, and pattern matching without pedagogical judgment risks recommending misaligned resources while reproducing biases embedded in training data \citep{Zhai2021Review, Holstein2021DesigningFair}. Equally important is the often overlooked role of learners’ cognitive and emotional resources, which must be efficiently allocated to support meaningful learning \cite{mehrabi2025uncovering, mehrabi2024ai}. Measurement-first systems typically treat students as vectors in a latent space and extend this logic across larger datasets, overlooking the fact that information extraction imposes costs measured in cognitive load, fatigue, and frustration. An adaptive loop that ignores attention as a bounded budget may continue to present diagnostically optimal items even as engagement declines, producing diminishing returns, and a loop that ignores frustration may mistake persistence for resilience while the learner approaches overload. By privileging measurement over experience, such systems make the diagnostic signal more precise even as its instructional yield becomes increasingly fragile \citep{conrad1972cognitive}.  
In this work, pedagogical expertise functions as the arbiter that ensures diagnostic insights trigger conceptual repair rather than superficial correction, while content selection prioritizes learning activities that are appropriate for all learners to mitigate bias \citep{Koedinger2013NewPotentials, Wang2021BigDataClustering}. The instructor-guided framework treats attention and affect as bounded resources; therefore, content recommendations are shaped jointly by evidence of conceptual states and the learner’s ability to engage with the additional cognitive demand of the educational resources. In other words, the framework not only considers which knowledge gap is pedagogically most meaningful to address but also which knowledge gap can be closed before the learner's cognitive resources are depleted. In practice, the system emulates expert educator decision-making to select individualized educational content resources that target diagnosed misconceptions while accounting for the learners' explicit limits of time, cognitive load, and emotional resilience \citep{Holstein2021DesigningFair, Mavrikis2015, Xie2019TrendsCompEdu}. The contributions of this framework are threefold. First, it outlines a normative approach in which effective adaptivity involves closing conceptual gaps while respecting learners’ limited cognitive resources and ensuring equitable support. Second, the framework uses Cognitive Diagnostic Models and a Q-matrix to identify which concepts are linked to each exam error and then assigns targeted learning resources accordingly. Third, the framework introduces a scalable, expert-validated pipeline that preserves pedagogical integrity in large and diverse learning contexts while providing open, customizable AI tools that instructors can adapt without advanced technical expertise.

\section{Literature Review}
\label{sec:litreview}

\subsection{From measurement-first adaptivity to instruction-centered design}
Early intelligent tutoring systems (ITS) established that fine-grained interaction traces—item responses, latencies, hint usage—could be transformed into moment-to-moment instructional decisions, yielding measurable learning gains and scalable personalization in authentic classrooms \citep{Anderson1985, vanlehn2006behavior, soloway1994learner}. As the field matured, however, innovation increasingly concentrated on \emph{diagnosis}: psychometric and machine-learning approaches refined latent trait estimation, reduced posterior uncertainty, and optimized assessment item selection using information-theoretic criteria \citep{de2011generalized, Rupp2008, Xie2019TrendsCompEdu}. While this measurement-first direction produced reliable improvements in predictive accuracy and assessment efficiency, reviews repeatedly note a disjunction between precise cognitive diagnosis and \emph{durable} conceptual change when subsequent instructional moves remain procedural, superficial, or poorly aligned with learners’ mental models \citep{Koedinger2013NewPotentials, Mavrikis2015, Holstein2021DesigningFair}. Risk is amplified when recommendation layers privilege convenient signals—popularity, brevity, easy-to-index topics—over conceptual suitability and cultural relevance; in such cases, short-term correctness can rise without restructuring underlying conceptions \citep{Holstein2021DesigningFair, mehrabi2024ai}.

Contemporary scholarship, therefore, is moving in a direction to reframe adaptivity as \emph{instruction-centered}. In other words, while diagnostic evidence is necessary, it is not sufficient unless assessments terminate in \emph{concept-level} remediation that learners can productively sustain. This is particularly critical given that learners have bounded attention and cognitive load, which when ignored often lead to cognitive fatigue, frustration, and disengagement \citep{DMello2012, conrad1972cognitive, Holstein2021DesigningFair}. The resulting research agenda shifts emphasis from forecasting responses to orchestrating concept repair that is meaningful to learners under resource limits, with diagnosis serving as a means rather than an end. In this view, adaptivity has two simultaneous obligations: (i) to identify which ideas or skills are most instructionally meaningful for a learner to address next, and (ii) to ensure that the timing, intensity, and form of the instructional support are feasible within learners’ cognitive and emotional resources.
Increasingly, the literature critiques adaptive algorithms and platforms that operate as open-loop systems, optimizing for information gain while postponing instructional action. When this happens, the responsibility to respond to a knowledge gap often falls on learners, who may not know how to address it, or on instructors, who may be unable to provide immediate feedback or support. By the time an appropriate intervention is offered, the opportunity for a well-timed conceptual response may have already passed \citep{Holstein2021DesigningFair}.


\subsection{ITS as an evidence-to-action architecture with instructor oversight}
The canonical ITS decomposition—\emph{Learner} model (knowledge, misconceptions), \emph{Domain} model (concept graph, Q-matrix), \emph{Tutoring} model (policies and scaffolds), \emph{Interface} model (presentation and affect awareness), and an \emph{Adaptive Engine}—persists because it provides an auditable bridge from evidence to action \citep{Woolf2010, vanlehn2006behavior, koedinger1997intelligent, Wenger1987}. Decades of work detail how learner models encode evolving mastery and misconceptions \citep{Conati2002, mehrabi2025uncovering}, how domain models structure prerequisite relations \citep{VanLehn2005}, and how tutoring policies select hints, examples, and practice to maximize learning opportunities rather than mere correctness \citep{Nkambou2010Advances}. Crucially, the decomposition includes \emph{policy hooks} for load- and affect-aware control: interface and tutoring layers can modulate pacing and representation in response to signs of confusion, overload, or frustration, thereby preventing diagnostically “optimal” yet instructionally counterproductive actions \citep{DMello2012, Holstein2021DesigningFair}. The architecture is thus not only a historical template; it is the locus at which measurement, pedagogy, and user experience are reconciled.

Within this architecture, the literature argues for \emph{instructor-orchestrated adaptivity}: teachers retain authority over the mapping from evidence to permissible actions by curating concept-indexed micro-interventions, auditing cultural fit, and defining acceptable evidence, while the system executes selection within those guardrails \citep{Aleven2016TeacherDashboard, Koedinger2013NewPotentials, Holstein2021DesigningFair, Mavrikis2015, Rodrigues2023LeveragingCV}. Large language models (LLMs) increasingly support scale—tagging resources, generating alternative explanations, proposing feedback variants—but are routed through the same auditable policies to limit preference-first drift and bias amplification \citep{Holstein2021DesigningFair, mehrabi2024ai}. The consensus in reviews is that such oversight preserves transparency and accountability while enabling automation where it demonstrably helps, aligning with longstanding ITS commitments to traceable “why this, why now?” justifications.

\subsection{Cognitive Diagnostic Models and Q-matrix validation as the instructional backbone}
Cognitive Diagnostic Models (CDMs) formalize a shift from item-level prediction to \emph{attribute-level} inference. A Q-matrix maps items to skills, and mastery vectors become the primary currency for instruction rather than opaque scores \citep{de2011generalized, Rupp2008}. Conjunctive families such as DINA align with prerequisite logic—success on an item presupposes a specific attribute set—while generalizations preserve interpretability, which is valued by instructors who must understand and contest model assumptions \citep{de2011generalized}. Beyond inference, the literature emphasizes that the Q-matrix is \emph{researchable}, meaning that it serves to: empirically validate and detect skill mastery misspecifications through discrimination checks, likelihood-based comparisons, and other methods. Thus the Q-matrix improves both classification fidelity and the downstream quality of instructional decisions that depend on those posteriors \citep{Rupp2008}. 

Studies connecting CDMs to resource selection report two practical advantages. First, mastery vectors are semantically aligned with curricular constructs, enabling direct indexing of resources to targeted attributes and prerequisites. Thereby, grounding content selection in pedagogy rather than topical proximity \citep{de2011generalized}. Second, explicit mappings make the system open to professional judgment, meaning that instructors can revise concept definitions and item–attribute links as relevant to their individual contexts or as domain understanding evolves. This adaptive ability preserves the auditability that ITS architectures prize \citep{Woolf2010, vanlehn2006behavior}. CDMs are effective as the “instructional backbone” not because they maximize predictive accuracy, but rather because they \emph{translate} evidence into concept-level levers that tutoring policies can reliably act upon.

\subsection{Micro-interventions under cognitive and affective budgets: representation, density, timing}
Instructional effectiveness depends not only on \emph{what} is taught, but on \emph{how} it is represented and \emph{how much} attentional and working-memory demand it imposes on learners. Cognitive Load Theory and multimedia learning principles yield actionable design guidelines that are characteristic of effective instruction, such as, minimizing extraneous processing, staging intrinsic load, and aligning representational form with the specific conceptual move being taught \citep{conrad1972cognitive}. Research on affect-aware tutoring further documents that learner affect (e.g., frustration, confusion, and fatigue) interacts with cognitive load to shape learning, and that ignoring learner affect and cognitive load often results in reduced learning or brittle, non-transferable learning gains \citep{DMello2012, Holstein2021DesigningFair}. Empirical studies of video-based learning at scale report advantages for concise, highly signposted explanations relative to long monologues, especially when the intervention deliberately \emph{switches representation} (symbolic $\leftrightarrow$ diagrammatic) to reframe stubborn misconceptions \citep{guo2014video, Zhai2021Review}. 

In engineering education, where conceptual understanding, mathematical formalisms, and modeling are braided, the constraints of learner affect and cognitive load are particularly salient.  Learners often require representational pivots rather than \emph{“more of the same”}, making the cost of poorly timed or overly dense material have an amplified negative impact \citep{Zhai2021Review, Holstein2021DesigningFair}. The cognitive load literature therefore, treats the alignment of time, redundancy, and difficulty as \emph{primary importance} considerations for micro-intervention design (e.g., short, skill-targeted videos (Content), worked examples with clear signaling, the use of multiple representations per concept, and explicit prerequisite metadata) to ensure that recommended actions are productive for robust long-term learning \citep{guo2014video, Mavrikis2015}. This body of work reframes engagement analytics such as dwell time, hint streaks, and latency spikes as inputs for setting instructionally meaningful constraints, rather than as retrospective dashboard metrics that describe engagement without informing real-time instructional decisions.


\subsection{Selection paradigms in practice: rule-based, similarity-driven, learner-centered}
Deployed systems typically balance or make trade-offs among transparency, continuity, and pedagogical precision, as design choices that strengthen one of these dimensions can constrain the others. Rule-based intervention selection attaches simple instructional policies to learner telemetry. For example, when a student makes two consecutive errors on a given concept, the system may automatically assign a worked example on that concept. Similarly, low dwell time on an instructional resource may prompt the system to deliver a shorter follow-up clip. These mechanisms offer operational simplicity and preserve instructor control. However, they also risk misalignment when the rules are not explicitly tied to the conceptual structure of the content or validated against diagnostic evidence \citep{Woolf2010, halvonik2020content, Holstein2021DesigningFair, Rodrigues2023Adaptive}.

Various research paradigms appear to converge on several safeguards: curation of diverse resource banks to widen representational coverage; implementation of subgroup-aware diagnostics to surface disparate impacts; and the requirement of auditable rationales for each recommendation so that educators can interrogate and refine decision rules \citep{Mavrikis2015, Holstein2021DesigningFair, mehrabi2024ai}. “Big data” clustering can diversify candidate pools and reduce duplication in large repositories \citep{Wang2021BigDataClustering}, but is most effective when embedded within concept-first and instructor-audited control loops. The overarching implication is that personalization earns instructional credibility when it remains anchored in concept repair, manages redundancy explicitly, and preserves professional oversight.

\subsection{Optimization strategies for constrained content selection: greedy, GRASP/$\varepsilon$-greedy, gradient-tuned scoring}
Selecting a \emph{set} of micro-interventions is combinatorial and time-bounded by classroom realities. When relevance, coverage, and redundancy penalties induce diminishing returns, greedy selection provides strong anytime performance with appealing interpretability. Randomized variants such as GRASP and $\varepsilon$-greedy inject controlled exploration to mitigate myopia under noisy diagnostics and heterogeneous resource banks \citep{resende2010greedy, Li2021Hyper}. Where data permit parameter learning, gradient-descent families—RMSprop and Adam in particular—tune trade-offs among length, alignment, and redundancy in underlying scoring functions; matrix factorization and related approaches help when repositories are large and sparse \citep{koren2009matrix, duchi2011adaptive, Kingma2015Adam}. Operational studies in education and scheduling corroborate the greedy algorithm’s responsiveness at scale while noting local-optima risks that motivate hybrid designs. For example, greedy scheduling on top of gradient-tuned scores; light exploration to discover underused, high-utility resources; and online data selection methods to adapt to distribution shift \citep{Min2016Greedy, Jiang2024AdaptiveDataOpt}. 

Importantly, recent reviews in technology-enhanced learning warn that optimization goals that maximize engagement or convenience alone can entrench bias or encourage redundancy \citep{Holstein2021DesigningFair, Rodrigues2023Adaptive}. The strongest reports frame intervention selection as an explicitly constrained problem that seeks to maximize instructionally grounded utility while enforcing budget limits (time and cognitive load), redundancy control, concept coverage, and transparency requirements that support teacher oversight \citep{Mavrikis2015, Holstein2021DesigningFair}. Within this framing, simple greedy methods remain attractive for their speed and explainability, while gradient-tuned scoring offers data-driven calibration of trade-offs; hybrid strategies balance the two, yielding systems that are operationally viable yet pedagogically principled \citep{resende2010greedy, Li2021Hyper, koren2009matrix, Kingma2015Adam, Wang2021BigDataClustering}.


\section{Methodology}
\subsection*{Conceptual Formulation and Objectives}

We formulate personalized instructional content (specifically, here instructional video-type of content) assignments as a constrained, multi-objective set-selection problem over a teacher-curated repository, with decisions acting on concept attributes (skills) rather than items or topics to keep “why this, why now?” auditable and instruction-centered \citep{Holstein2021DesigningFair, Koedinger2013NewPotentials, Rodrigues2023Adaptive}. Let $N$ be learners, $M$ content, and $K$ skills defined by an instructor-authored Q-matrix; for learner $i$, the mastery vector $\mathbf{S}_i=(S_{i1},\ldots,S_{iK})\in\{0,1\}^K$ induces non-mastery $\mathbf{U}_i=\mathbf{1}-\mathbf{S}_i$, while each content $j$ has coverage $\mathbf{C}_j=(C_{j1},\ldots,C_{jK})\in\{0,1\}^K$ and length $V_j>0$; the binary decision $x_{ij}\in\{0,1\}$ indicates whether content $j$ is assigned to learner $i$.

The objective structure balances adequacy and parsimony. We reward concept repair by aligning coverage with non-mastery using $Z_1=\sum_{i,j,k}U_{ik}C_{jk}x_{ij}$ and discourage attention waste via $Z_2=\sum_{i,j}x_{ij}+\epsilon\sum_{i,j}V_jx_{ij}$ (with $\epsilon>0$ scaling time relative to count) \citep{Xie2019TrendsCompEdu, Zhai2021Review}. A standard scalarization combines these goals,
\[
Z=\alpha Z_1-\beta Z_2,
\]
and we adopt a sufficientarian stance in which adequacy is the floor: either enforce gap-closure as a hard requirement and then minimize burden (lexicographic or $\varepsilon$-constraint) or choose $\alpha/\beta$ large enough that any improvement in $Z_1$ dominates plausible changes in $Z_2$ \citep{Holstein2021DesigningFair, Rodrigues2023Adaptive}. This keeps fairness a guarantee rather than a post-hoc metric.

\subsection*{Constraints and Pedagogical Safeguards}

For learner $i$, the feasible assignment vector $x_i=(x_{i1},\dots,x_{iM})$ belongs to $\mathcal{X}_i=\{x_i\in\{0,1\}^M:\text{(1)--(6) hold}\}$. We list the constraints as numbered commitments and then describe the fallback policy in the same narrative style.

(1) \emph{Gap-closure (adequacy).} Every diagnosed skill gap must be addressed: for each skill $k$, the coverage condition is $\sum_j C_{jk}\,x_{ij}\ge U_{ik}-\xi_{ik}$ with slack $\xi_{ik}\ge 0$ used only when the repository is insufficient; slacks are heavily penalized in the objective and always reported to instructors. This renders adequacy a guarantee (or a surfaced deficiency) rather than a post hoc metric.

(2) \emph{Cognitive/temporal budgets (feasibility).} Attention is finite; we bound total duration and slate size per learner by enforcing $\sum_j V_j\,x_{ij}\le T_{\max,i}$ and $\sum_j x_{ij}\le B_i$. These limits reflect classroom realities and cognitive-load considerations for micro-interventions.

(3) \emph{Ability-aligned difficulty windows (appropriateness).} Each content $j$ has a difficulty index $d_j$ (from instructor rubrics or historical use), and each learner $i$ has a readiness window $[\ell_i,u_i]$ derived from ability and course policy; we require that any assigned item respects the window, i.e., $x_{ij}=1\Rightarrow d_j\in[\ell_i,u_i]$ (implemented via a standard big-$M$ linearization). This prevents “knowledge-optimal but cognitively inappropriate” slates.

Diagnostic signals inform, but do not dominate, the formulation: ability $\theta_i$ (from IRT/3PL) governs \emph{readiness}—the difficulty windows and time budgets a learner can productively sustain—while attribute-level mastery $S_{ik}$ (from DINA/CDMs) encodes \emph{responsibility}—the specific concepts we are obligated to repair; a single scalar cannot express both readiness and localization, so 3PL/IRT calibrates feasibility terms (e.g., difficulty alignment, $T_{\max}$, slate size) whereas DINA/CDMs supply the gap vector $U_{ik}=1-S_{ik}$ that anchors the coverage objective $Z_1$ and constraint (C1), ensuring selections act on causes of error rather than topical similarity \citep{de2011generalized, Rupp2008, vanlehn2011relative, Holstein2021DesigningFair}. Two design invariants follow: first, \emph{concept-first targeting with teacher governance}—the Q-matrix and repository metadata $\{\mathbf{C}_j,V_j\}$ are instructor-authored, enabling transparent overrides and routine Q-matrix/metadata repair when misfit appears \citep{Holstein2021DesigningFair}; second, \emph{budget awareness}—attention and redundancy carry explicit costs, aligned with cognitive-load guidance for micro-interventions \citep{Zhai2021Review}. The optimizer choice (greedy in austerity regimes, gradient-tuned in abundance) is orthogonal to these roles and selected by deployment constraints, while objectives and safeguards remain unchanged \citep{resende2010greedy, rodrigues2023equity}.

(4) \emph{Prerequisite coherence (no leaps).} With a prerequisite graph $A$ where $A_{k\to k'}=1$ indicates $k$ precedes $k'$, we forbid assigning content that targets $k'$ unless $k$ is already mastered or simultaneously remediated; a sufficient linear condition is $\sum_j C_{jk'}\,x_{ij}\le S_{ik}+\sum_j C_{jk}\,x_{ij}$ for all prerequisite pairs $(k\!\to\!k')$.

(5) \emph{Anti-redundancy and representational diversity (attention stewardship).} To reduce near-duplicate recommendations, for any pair with high similarity (e.g., $(j,\ell)\in\mathcal{R}_\tau$ when $\mathrm{sim}(j,\ell)\ge\tau$), we enforce $x_{ij}+x_{i\ell}\le 1$. To encourage conceptual reframing, we also require at least $\delta$ distinct representational forms per slate: with $M_{jr}\in\{0,1\}$ indicating that content $j$ uses representation $r$ and binary auxiliaries $z_{ir}$, we use $z_{ir}\le\sum_j M_{jr}x_{ij}$ and $\sum_r z_{ir}\ge \delta$.

(6) \emph{Integrality and auditability.} Assignment and diversity indicators are integral ($x_{ij}\in\{0,1\}$, $z_{ir}\in\{0,1\}$) and slacks nonnegative ($\xi_{ik}\ge 0$); every recommendation carries a trace back to $(U_{ik},\theta_i)$ and repository metadata $(C_{jk},V_j,d_j,M_{jr})$, enabling routine Q-matrix and metadata repair when misfit persists.

\emph{Content-level fallback (difficulty-aware selection under scarcity).} Difficulty is organized on an ordered scale $\text{basic}<\text{medium}<\text{hard}$. Each content $j$ has a level $D_j$ on this scale, and each learner $i$ has a preferred level $P_i$ implied by readiness (e.g., from $\theta_i$) and course policy. Selection proceeds with a distance-aware score that balances coverage against burden and difficulty deviation: $F_{ij}=\bigl|S_i\cap C_j\bigr|-\bigl(\epsilon\,L_j+\omega\cdot\mathrm{dist}(D_j,P_i)\bigr)$, where $\mathrm{dist}$ is the level gap (0, 1, or 2), $\epsilon>0$ scales time, and $\omega>0$ penalizes misalignment. The policy is: attempt to satisfy (1)–(6) using content with $D_j=P_i$; if adequacy remains unmet within budgets, admit one-level deviations (i.e., $\mathrm{dist}=1$) with the additive penalty $\omega$; only if coverage is still infeasible within (2)–(4) do we admit two-level deviations with penalty $2\omega$. At each admission tier we greedily or globally select among admissible items by maximizing $F_{ij}$ subject to (1)–(6), update uncovered skills and budgets, and continue until all $U_{ik}$ are covered or budgets are exhausted. Any residual unmet skills are recorded as slacks $\xi_{ik}>0$ and surfaced for instructor action. This distance-penalized fallback keeps the slate cognitively appropriate whenever possible, degrades gracefully under scarcity, and remains auditable. This means that deviations from $P_i$ are explicit, bounded, and justified by repository constraints rather than by opaque heuristics.
\subsection*{Formal Guarantees (Sketch)}

For each learner $i$, an assignment vector $x_i\in\{0,1\}^M$ is admissible only if it satisfies constraints (1)–(6), i.e., $x_i\in\mathcal{X}_i=\{x_i:\text{(1)–(6)}\}$, and admissible slates are ranked by the scalarized objective $Z(x)=\alpha Z_1(x)-\beta Z_2(x)$, which rewards concept repair and penalizes attention cost. The constraints define what is pedagogically acceptable; the objective orders acceptable solutions.

\textbf{Adequacy, budgets, and coherence.} If (1) holds without slack ($\xi_{ik}=0$), then every diagnosed gap is covered: whenever $U_{ik}=1$ there exists at least one assigned content $j$ with $C_{jk}=1$ (adequacy). If (2) holds, total duration and slate size are bounded by $T_{\max,i}$ and $B_i$ (feasibility), and if (3)–(4) hold, all assigned items respect the learner’s difficulty window $[\ell_i,u_i]$ and prerequisites (appropriateness and coherence).

\textbf{Proposition 4 (Existence/diagnostic of infeasibility).} Strict adequacy (no slacks) is possible if and only if the repository covers every required skill for that learner:
\[
\forall (i,k)\ \text{with}\ U_{ik}=1,\ \exists j\ \text{such that}\ C_{jk}=1.
\]
If this condition fails for some $(i,k)$, then (1) can only be met with $\xi_{ik}>0$, which we \emph{penalize and report}, yielding a certificate of repository insufficiency at that learner–skill pair. Necessity is immediate; sufficiency holds by selecting at least one covering item per uncovered $(i,k)$.

\textbf{Complexity of exact solution.} Solving the exact integer program is NP-hard. Even with only (1) and a single budget from (2), the problem reduces to maximum-coverage/knapsack variants; adding difficulty windows, prerequisites, and anti-redundancy/diversity (3)–(5) does not make it easier. Hence, we rely on principled approximations rather than exact IP at scale.

\textbf{Greedy guarantees under diminishing returns.} Using a capped-coverage surrogate that counts each learner–skill at most once and keeping (2) as a knapsack-style budget while implementing (5) via pairwise exclusion or a concave redundancy penalty yields a monotone submodular objective. The classic greedy rule is then $(1-1/e)$–approximate under a single knapsack and $1/2$–approximate under typical diversity constraints—matching the regime induced by time budgets and anti-redundancy.

\textbf{Gradient-based relaxation and rounding.} Relaxing $x_{ij}\in\{0,1\}$ to $x_{ij}\in[0,1]$ and optimizing a smooth surrogate $\hat Z=\alpha\hat Z_1-\beta\hat Z_2$ with differentiable barriers for (2)–(4) gives projected gradient descent that converges to the optimum in convex cases (or to a first-order stationary point otherwise). Thresholding recovers an integer slate preserving (2)–(4) and only misses adequacy where content is genuinely absent—already surfaced by slacks in (1).

\textbf{Fairness as control, not a metric.} Treating adequacy as the floor can be posed either as an $\varepsilon$–constraint/lexicographic model (enforce (1), then minimize burden) or as a weighted sum with $\alpha/\beta$ large enough to grant lexical priority to coverage; under reasonable frontiers, these yield the same Pareto–efficient choices.

\textbf{Stability to small changes.} Adding new content to the repository can only improve capped coverage (monotonicity), and flipping a single $U_{ik}$ from 0 to 1 changes the optimal capped objective in a controlled, one-step way, up to the budgets in (2). Thus, minor metadata edits or incremental content growth will not destabilize assignments.

\textbf{Richness and diversity (operational definitions and regime law).} All guarantees operate on the \emph{admissible pool} that remains after applying feasibility safeguards: for learner $i$, let $\mathcal{A}_i=\{j:\text{content $j$ respects (2)–(4) for learner $i$}\}$ and remove near-duplicates per (5) to obtain a nonredundant pool $\tilde{\mathcal{A}}_i$. We use \emph{richness} to denote the effective variety of well-mapped, difficulty-appropriate options per uncovered skill after these filters; an operational composite is $\rho_i=w_1 b_i+w_2 \tilde c_i+w_3 \tilde H_i+w_4 \tilde\sigma_i$, where $b_i$ is breadth (fraction of required skills in $U_i$ that have at least one admissible item), $\tilde c_i$ is the median number of admissible, nonredundant items per required skill (over $\tilde{\mathcal{A}}_i$), $\tilde H_i$ is normalized representation entropy across forms (e.g., symbolic/diagrammatic/narrative), and $\tilde\sigma_i$ is normalized difficulty spread within $[\ell_i,u_i]$; each term is scaled to $[0,1]$. \emph{Diversity} is the representational variety and anti-redundancy enforced by (5), including a minimum distinct-forms target $\delta$ and exclusion of high-similarity pairs $\mathcal{R}_\tau$. The regime law follows: in “austerity” (low $\rho_i$) and tight latency/compute, greedy is preferred for its speed and approximation guarantees; in “abundance” (high $\rho_i$) and looser budgets, gradient-based coordination pays off; between these extremes, hybrids that initialize greedily over $\tilde{\mathcal{A}}_i$ and refine with gradient steps align with (1)–(6) and preserve auditability. Cohort-level richness can be summarized by the median (or quantiles) of $\{\rho_i\}$ to choose the optimizer policy.

\medskip
These statements justify (i) \emph{why} exact IP is not the baseline (Theorem~1), (ii) \emph{when} GH is principled (Theorem~2), (iii) \emph{how} GD’s relaxation behaves (Theorem~3), and (iv) \emph{how} our equity stance maps to standard multiobjective control (Lemma~1)—while preserving the core adequacy and budget guarantees (Propositions~1–3). We separate diagnostics by function: $\theta_i$ governs readiness (difficulty windows, time budgets) while $S_{ik}$ governs responsibility (which gaps to repair). Because a single scalar cannot express both readiness and localization, 3PL/IRT informs feasibility constraints and DINA/CDMs supply the gap vector $U_{ik}$ that drives $Z_1$ and (C1), yielding cognitively appropriate yet concept-adequate slates \citep{de2011generalized, Rupp2008, Holstein2021DesigningFair}.

\subsection*{Algorithmic Properties and Optimization Regimes}

\textbf{Theorem 1 (Computational hardness).} The exact integer program is NP-hard.   With only (C1) and a single global budget $\sum_j V_j x_{ij}\le T_{\max,i}$, maximizing $Z_1$ reduces to \emph{Maximum Coverage with a Knapsack} (MCKP); setting $V_j\equiv 1$ reduces to \emph{Maximum Coverage}/\emph{Set Cover}—both NP-hard. Adding (C3)–(C5) does not ease hardness.

\textbf{Theorem 2 (Greedy approximation under diminishing returns).} Define a \emph{capped} coverage surrogate
\[
\tilde Z_1(x)\;=\;\sum_{i,k}U_{ik}\,\min\!\Big\{1,\ \sum_j C_{jk}x_{ij}\Big\},
\]
and let the budget be knapsack-style $\sum_j V_j x_{ij}\le T_{\max,i}$ (per learner), with (C5) implemented as either (i) pairwise exclusion of near-duplicates or (ii) a concave redundancy penalty. Then $\tilde Z_1$ is \emph{monotone submodular} in the chosen set of content; the classic greedy rule yields a $(1-1/e)$-approximation to $\max \tilde Z_1$ under a single knapsack constraint, and a $1/2$-approximation under a matroid intersection (e.g., diversity) constraint.   Capped coverage and concave gains induce diminishing returns; apply standard results for submodular maximization with knapsack/matroid constraints.

\textbf{Theorem 3 (Convergence of gradient-based relaxation).} Consider the continuous relaxation $x_{ij}\in[0,1]$ with a smooth surrogate objective
\[
\hat Z(x)\;=\;\alpha\,\hat Z_1(x)\;-\;\beta\,\hat Z_2(x)
\]
in which the cap $\min\{1,\cdot\}$ is replaced by a smooth concave approximation and penalties for (C2)–(C4) are added via differentiable barriers. If $\hat Z$ is $L$-smooth and (weakly) convex, projected gradient descent with step sizes $\eta_t\le 1/L$ converges at rate $O(1/t)$ to the global optimum; if nonconvex but $L$-smooth, it converges to a first-order stationary point. Rounding by thresholding recovers a feasible integer solution that preserves (C2)–(C4) and violates (C1) only where the repository is insufficient (flagged by slacks).   Standard convex optimization arguments for smooth objectives; for nonconvex surrogates, apply descent lemmas and projected gradient convergence to stationary points.

\textbf{Lemma 1 (Equivalence of fairness stances).} Let the \emph{$\varepsilon$-constraint} formulation enforce (C1) exactly (or with $\xi_{ik}$ penalized by a coefficient $\gamma\!\to\!\infty$), optimizing parsimony thereafter. Then the weighted-sum formulation $Z=\alpha Z_1-\beta Z_2$ with $\alpha/\beta$ sufficiently large is equivalent in the sense of producing the same Pareto-efficient solutions (up to ties).   Weighted-sum and $\varepsilon$-constraint are equivalent for convex frontiers; with a discrete frontier, taking $\alpha/\beta$ larger than the maximum possible change in $Z_2$ induced by one unit of $Z_1$ secures lexicographic priority to adequacy.

\textbf{Lemma 2 (Monotonicity and stability).} If $C$ is augmented by an additional content or if $U$ flips a single $(i,k)$ from $0$ to $1$, the optimal value of $\tilde Z_1$ is nondecreasing, and changes are $1$-Lipschitz in the Hamming distance on $(U,C)$ up to the imposed budgets.   Monotonicity follows from submodularity; Lipschitz bounds follow from the cap and unit changes in attainable coverage.

\paragraph{Design law (switching heuristic).} Let $\rho$ denote repository richness (diversity $\times$ granularity) and $\lambda$ the latency/compute budget. Use GH when $(\rho,\lambda)$ are below a regime threshold $(\rho^\star,\lambda^\star)$ (austerity); prefer GD when above (abundance); interpolate with a hybrid otherwise. This is consistent with Theorem~2 (greedy’s guarantees under diminishing returns) and Theorem~3 (benefits of global tuning when time permits).

\medskip
These statements justify (i) \emph{why} exact IP is not the baseline (Theorem~1), (ii) \emph{when} GH is principled (Theorem~2), (iii) \emph{how} GD’s relaxation behaves (Theorem~3), and (iv) \emph{how} our equity stance maps to standard multiobjective control (Lemma~1)—while preserving the core adequacy and budget guarantees (Propositions~1–3).


\subsection*{Algorithmic Implementations, Evaluation Design, and Metrics}

We implement two optimization approaches that respect constraints (1)–(6) but differ in how they trade computation for global coordination. Throughout, candidate assignments for learner $i$ are drawn from the nonredundant, admissible pool $\tilde{\mathcal{A}}_i$ obtained after enforcing time/size budgets (2), difficulty windows (3), prerequisite coherence (4), and anti-duplication filters in (5). Richness $\rho_i$ (effective variety after filters) guides which approach is preferable in practice (cf.\ the regime law).

\emph{Greedy Heuristic (GH), sequential and myopic but fast.} The greedy policy constructs a slate $V_i$ iteratively. At iteration $t$, let the still-uncovered skill set be $U_i^{(t)}=\{k:U_{ik}=1\ \text{and}\ k\ \text{not yet covered by}\ V_i^{(t)}\}$. We score each admissible content $j\in\tilde{\mathcal{A}}_i$ by a distance-aware, redundancy-aware net-gain function that balances coverage against burden and misalignment,
\[
F_{ij}^{(t)} \;=\; \bigl|U_i^{(t)}\cap C_j\bigr|\;-\;\bigl(\,\epsilon\,L_j\;+\;\omega\,\mathrm{dist}(D_j,P_i)\;+\;\gamma\,\mathrm{overlap}_j^{(t)}\,\bigr),
\]
where $L_j$ is length, $\mathrm{dist}(D_j,P_i)\in\{0,1,2\}$ measures the level gap between the content’s difficulty and the learner’s preferred level (cf.\ fallback), and $\mathrm{overlap}_j^{(t)}=\bigl|(\cup_{\ell\in V_i^{(t)}}C_\ell)\cap C_j\bigr|$ penalizes skill redundancy; $\epsilon,\omega,\gamma>0$ set the burden, misalignment, and redundancy penalties. The algorithm chooses $j^\star=\arg\max_{j\in\tilde{\mathcal{A}}_i}F_{ij}^{(t)}$ subject to (2)–(5) remaining feasible if $j^\star$ is added; then it updates $V_i^{(t+1)}=V_i^{(t)}\cup\{j^\star\}$, refreshes $U_i^{(t+1)}$, and continues until $U_i^{(t)}=\emptyset$ or either time $T_{\max,i}$ or cardinality $B_i$ is reached. When no $j$ with $\mathrm{dist}=0$ can complete coverage within budgets, the distance-penalized fallback admits $\mathrm{dist}=1$ and then $\mathrm{dist}=2$ items (each with additive penalty $\omega$ per level) while keeping (1)–(6) intact. This selection realizes the capped-coverage, diminishing-returns structure required for submodular guarantees and runs in $O(|\tilde{\mathcal{A}}_i|\log|\tilde{\mathcal{A}}_i|)$ per iteration due to sorting by $F_{ij}^{(t)}$, leading to $O(M\log M)$ per learner in typical classroom pools \citep{resende2010greedy, liu2017design}. Detail steps of this algorithm is available from the Algorithm \ref{alg:GreedyHeuristic}.

\emph{Gradient-based relaxation (GD), holistic and tunable but iterative.} To coordinate trade-offs across many viable alternatives (high $\rho_i$), we relax $x_{ij}\in\{0,1\}$ to $x_{ij}\in[0,1]$ and minimize a smooth, penalized loss that mirrors our instruction-centered objectives and safeguards. Let the smooth, capped-coverage surrogate be $\hat Z_1(x)=\sum_{k}U_{ik}\,\sigma\!\left(\sum_{j}C_{jk}x_{ij}\right)$, where $\sigma(z)=1-e^{-\tau z}$ (concave, $0\!\le\!\sigma\!\le\!1$, parameter $\tau>0$). Let $\hat Z_2(x)=\sum_{j}x_{ij}+\epsilon\sum_{j}L_jx_{ij}$. We then optimize
\[
\min_{x_{i\cdot}\in[0,1]^M} \;\; \mathcal{L}_i(x)\;=\;-\alpha\,\hat Z_1(x)\;+\;\beta\,\hat Z_2(x)\;+\;\lambda_{\text{time}}\,[\![\sum_j L_jx_{ij}-T_{\max,i}]\!]_+^2\;+\;\lambda_{\text{card}}\,[\![\sum_j x_{ij}-B_i]\!]_+^2
\]
\[
+\;\lambda_{\text{diff}}\sum_j \phi_{\ell_i,u_i}(d_j)\,x_{ij}\;+\;\lambda_{\text{pre}}\sum_{k\to k'}\Big[\![\sum_j C_{jk'}x_{ij}-S_{ik}-\sum_j C_{jk}x_{ij}]\!]\_+\Big]^2\;+\;\lambda_{\text{div}}\sum_{(j,\ell)\in\mathcal{R}_\tau} x_{ij}x_{i\ell},
\]
where $[\![z]\!]_+=\max(0,z)$, $\phi_{\ell_i,u_i}(d_j)$ is a smooth penalty for difficulty violations relative to the window $[\ell_i,u_i]$, the prerequisite term enforces (4) as a soft barrier, and the diversity term discourages near-duplicates per (5). Projected gradient descent updates $x_{ij}\leftarrow\Pi_{[0,1]}(x_{ij}-\eta\,\partial\mathcal{L}_i/\partial x_{ij})$ with stepsize $\eta>0$ and converges to the global optimum when $\mathcal{L}_i$ is convex (or a stationary point when nonconvex); thresholding at $\tau^\star\in(0,1)$ recovers a binary slate that preserves (2)–(4) and only fails adequacy where the repository is truly insufficient (already flagged by (1)’s slacks). The per-iteration cost is $O(MK)$ for gradient accumulation over coverage and prerequisite terms, yielding $O(NMK)$ overall across learners \citep{boyd2004convex, rodrigues2023equity}. Detail steps of this algorithm is available from the Algorithm Box \ref{alg:GradientDescent}.

\emph{Evaluation design (simulation and field).} We probe both approaches under controlled and authentic conditions. In simulation, a CAT engine generates responses that estimate ability $\theta_i$ via 3PL and mastery $S_{ik}$ via DINA, producing the dual diagnostics that instantiate readiness (difficulty windows, budgets) and responsibility (gap vectors) for slate construction; content are parameterized by $(C_{jk},L_j,d_j,D_j)$, with representation tags for diversity. In the field study, we apply the same pipeline to pre-test data from 1{,}204 students in an introductory physics course, using instructor-curated content and fixed-form assessments; item parameters for 3PL/DINA are estimated from historical cohorts, and the same constraints (1)–(6) govern selection. Figure~\ref{fig:adaptive_workflow} summarizes the end-to-end flow from responses to assignments.

\emph{Metrics and interpretation.} We evaluate sufficiency and stewardship using four complementary measures. \emph{Satisfactory Rate (SR)} is the fraction of learners who achieve full concept adequacy under (1), i.e., those for whom every $U_{ik}=1$ is covered by the assigned slate; operationally, $\mathrm{SR}=\frac{1}{N}\sum_i \mathbf{1}\{\forall k:U_{ik}=1\Rightarrow \sum_j C_{jk}x_{ij}\ge 1\}$. \emph{Gain Decay (GD)} quantifies diminishing returns due to redundancy as the marginal coverage per additional minute (or item) falls across the slate; we compute GD as the normalized drop in coverage gain between successive selections (lower is better, indicating temperance). \emph{Utility (U)} measures concept coverage per unit attention, $U=\frac{\sum_i\sum_k U_{ik}\,\mathbf{1}\{\sum_j C_{jk}x_{ij}\ge 1\}}{\sum_i\sum_j L_jx_{ij}}$, capturing parsimony. \emph{Total Penalty (TP)} aggregates equity- and feasibility-relevant costs—overcoverage beyond needs, unused long items suggested by the optimizer, difficulty misalignments, and any prerequisite or time slacks—weighted by instructor policy. Reported together, SR certifies adequacy (the floor), U captures attention stewardship, GD diagnoses redundancy creep, and TP reflects policy-relevant frictions. These metrics permit apples-to-apples comparisons between GH (fast, explainable choices under low richness) and GD (globally coordinated choices under high richness) while remaining aligned with the constraints and fairness stance encoded in (1)–(6), the fallback policy, and the regime law \citep{Holstein2021DesigningFair, Zhai2021Review, resende2010greedy, boyd2004convex}.

\section{Results}

\subsection{Study Setting}

\subsubsection*{Study Setting: Data and model parameters flow}
\begin{figure}[h]
\centering
\begin{tikzpicture}[
    node distance=8mm and 4mm,
    every node/.style={draw, rectangle, align=center, rounded corners, minimum height=2em},
    input/.style={fill=blue!20},
    process/.style={fill=green!20},
    process2/.style={fill=red!20},
    output/.style={fill=orange!20},
    line/.style={-Latex, thick},
    zigzagarrow/.style={
        -Latex,
        thick,
        decorate,
        decoration={
            zigzag,
            amplitude=0.5mm,
            segment length=2mm,
            post=lineto,
            post length=2pt
        }
    }
]
\node (studentdata) [input] {Student \\Responses \\in non-CAT Exam};
\node (videodata) [input, left=of studentdata] {Content Data \\ (\( C_{jk} \), \( L_j \), \( M \times K \))};
\node (cat) [process, right=of studentdata] {Student \\Responses \\ in CAT Exam};
\node (Itemparameter) [input, right=of cat, xshift=2cm, yshift=0cm] {Item Parameters \\from simulation};

\node (ability) [process, below=of cat, xshift=-4cm, yshift=-1cm] {Ability by 3PL \\ (\( \theta_i \))};
\node (dina) [process2,  right=of ability, xshift=2cm] {DINA Model\\Skill Mastery\\(\( S_{ik} \), \( N \times K \))};

\node (ipmodel) [output, below=of dina, xshift=-2cm, yshift=-1cm] {IP Model (GH or GD) \\ Objective: Min. Watch Time \\ Constraints: Skill Coverage};
\node (output) [output, right=of ipmodel, xshift=1cm] {Assignments \\ (\( x_{ij} \), \( N \times M \))};

\draw[line, dashed] (Itemparameter.west) -- (cat.east) node[midway, below, font=\small] {\( a_j, b_j, c_j\)};
\draw[line, dashed] (cat) -- (ability)  node[midway, below, font=\small] {\( Y_{ij} \)};
\draw[line, dashed] (cat) -- (dina)  node[midway, right, font=\small] {\( Y_{ij} \)};
\draw[zigzagarrow] (studentdata.south) -- (ability)  node[midway, left, font=\small] {\( Y_{ij} \)};
\draw[zigzagarrow] (studentdata.south) -- (dina)  node[midway, right, font=\small] {\( Y_{ij} \)};
\draw[line] (dina.south) -- (ipmodel.north) node[midway, right, font=\small] {\( S_{ik} \)};
\draw[line] (ability.south) -- (ipmodel.north) node[midway, left, font=\small]  {\( \theta_i, a_j, b_j, c_j \)};
\draw[line] (videodata.south) |- (ipmodel.west) node[pos=0.25, left, font=\small] {\( C_{jk}, L_j \)};
\draw[line] (ipmodel) -- (output) node[midway, below, font=\small] {\( x_{ij} \)};
\end{tikzpicture}
\caption{Integrated adaptive learning workflow showing real-data (zigzag) and simulated (dashed) pathways for CAT and assignment optimization}
\label{fig:adaptive_workflow}
\end{figure}
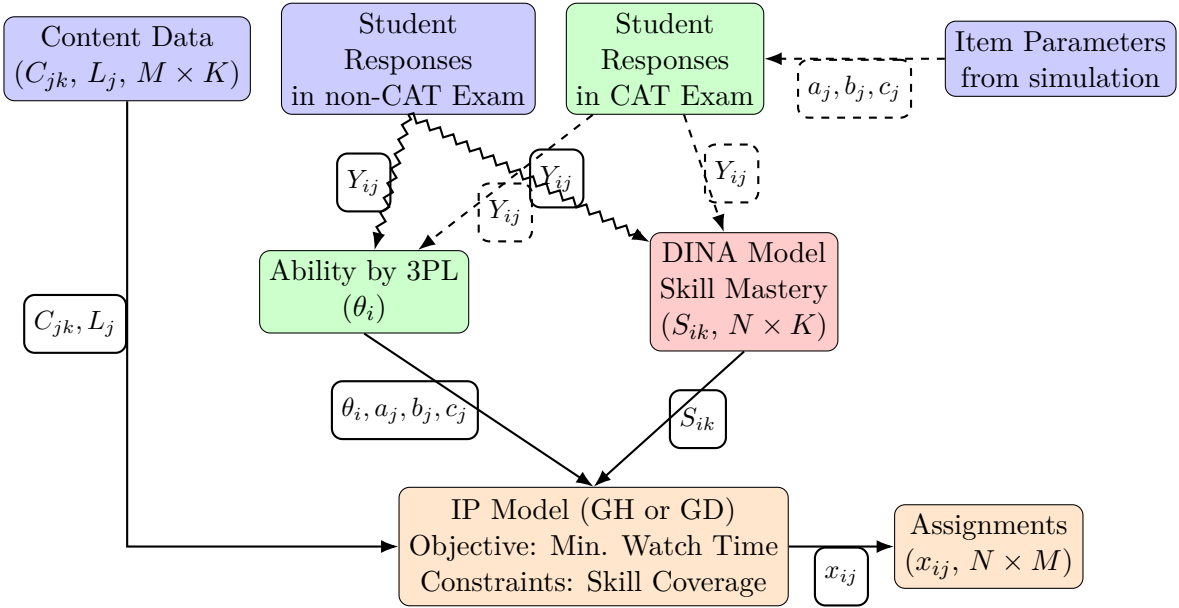

Our proposed evaluation framework, depicted in Figure \ref{fig:adaptive_workflow}, operationalizes a comprehensive and analytically rigorous dual-pathway diagnostic architecture that simultaneously captures students’ global proficiency and fine-grained skill mastery, directly addressing a pervasive limitation in conventional adaptive learning technologies which often rely on a singular dimension of learner information. The framework begins by ingesting raw student response data, which serves as the foundational input to two parallel but complementary diagnostic pipelines. This architectural decision is grounded in cognitive and educational psychology research indicating that effective personalized instruction requires both macro-level assessments of overall student ability and micro-level identification of discrete skill deficits—two qualitatively different insights that cannot be derived from a single diagnostic methodology. In the first diagnostic pathway, student responses are processed through a Computerized Adaptive Testing (CAT) engine driven by the 3-Parameter Logistic (3PL) Item Response Theory (IRT) model, which estimates student ability as a continuous latent variable \(\theta_i\). The 3PL model is expressed as \(P(y_{ni}=1|\theta_n) = c_i + (1 - c_i)/(1 + \exp(-a_i(\theta_n - b_i)))\), where each student's probability of correctly answering an item is modeled based on item parameters for discrimination (\(a_i\)), difficulty (\(b_i\)), and guessing (\(c_i\)). CAT dynamically selects items to maximize the information gain relative to the current estimate of \(\theta_i\), producing efficient, individualized assessments that converge on high-confidence ability estimates using a stopping criterion of either a standard error threshold of 0.2 or a maximum of 30 items. This ability measure informs the broader context of a student’s readiness to engage with content of varying complexity and rigor. In tandem, the second pathway employs the Deterministic Input, Noisy “And” (DINA) model, a Cognitive Diagnostic Model (CDM) that operates under a fundamentally different logic. Rather than estimating a single continuous trait, the DINA model determines a binary skill mastery profile for each student, based on whether a student possesses all the requisite skills needed to answer a given item correctly. The model is given by \(P(y_{ni} = 1 | \Theta_n) = (1 - s_i)^{\xi_{ni}} \cdot g_i^{1 - \xi_{ni}}\), where \(\xi_{ni} = \prod_{j=1}^{S} \Theta_{ns}^{S_{ij}}\), representing the conjunctive ("AND") assumption that all required skills must be mastered for success, and where \(s_i\) and \(g_i\) model slipping and guessing behavior respectively. The output is a binary matrix \(S_{ik}\) capturing student-level mastery across multiple latent skills, enabling precise identification of instructional gaps (Figure \ref{fig:adaptive_workflow}).

These two diagnostic outputs---continuous ability from the 3PL-based CAT and discrete skill mastery from DINA---are then synthesized within a unified optimization engine tasked with generating individualized instructional content assignments \(x_{ij}\) (Figure \ref{fig:adaptive_workflow}. Algorithmic details in Appendix).  

Integrating fundamentally different data types---continuous \(\theta_i\) and binary \(S_{ik}\)---into a coherent optimization model requires sophisticated mathematical treatment. The engine must reconcile these heterogeneous representations to achieve several competing instructional objectives.  

First, it aims to \textbf{minimize the total duration} of assigned content, thereby reducing students' cognitive load. Second, it seeks to \textbf{maximize the coverage of unmastered skills}, ensuring that each student's learning plan emphasizes targeted remediation. Finally, it strives to \textbf{align the difficulty level} of content with the student's global ability estimate, maintaining both motivation and instructional alignment (Figure \ref{fig:adaptive_workflow}).  

The instructional content itself is represented by a matrix \(C_{jk}\), indicating which skills are covered by each content, and a vector \(L_j\), capturing the corresponding content lengths. Together, these components define the optimization’s decision variables and constraints, forming the mathematical backbone of personalized instructional assignment (Figure \ref{fig:adaptive_workflow}).  


In simulation environments, the full diagnostic-optimization loop enables comprehensive benchmarking and sensitivity analyses, wherein parameters such as the number of available content, the distribution of student skill profiles, and ability levels can be systematically varied to evaluate algorithmic robustness and generalizability. CAT-generated responses emulate realistic student behavior under adaptive testing conditions, and the DINA model provides granular skill diagnoses that traditional test scores fail to uncover. For practical deployment, the same end-to-end workflow is applied to authentic student data, thereby enabling direct validation of its real-world efficacy. The consistent application of psychometric models and optimization algorithms across both simulation and operational settings allows for rigorous, comparative evaluations and enhances the external validity of simulation-derived insights. The entire framework is carefully designed to emulate the cognitive processes of skilled educators, who intuitively weigh both overall student performance and individual skill strengths or weaknesses when making instructional decisions. By formalizing this expert reasoning into computational procedures, the framework enables scalable, data-driven personalization at a level of pedagogical sophistication that traditional adaptive systems, which rely solely on item-level response correctness or a single trait score, cannot achieve. This integrated approach, grounded in psychometric theory, cognitive diagnostics, and optimization science, represents a significant advancement in the design of adaptive instructional systems capable of delivering nuanced and equitable educational interventions at scale. Detailed formulations of the 3PL and DINA models, along with performance metrics, are provided in Appendix A.

\subsubsection*{Study Setting: Simulating response data and validation}
Our comprehensive simulation study was designed to rigorously evaluate the performance of the proposed optimization framework under controlled conditions, incorporating systematically varied parameters to assess robustness and generalizability. For the simulation study, we designed a comprehensive evaluation with 1,000 synthetic students, 60 assessment items, and 5 latent skills organized into 3 content areas (20 items each), following established cognitive diagnostic methodologies \citep{wang2021improvement}. A Q-matrix $S_{ij} \in \{0,1\}^{I \times S}$ defines item-skill relationships, where $S_{ij} = 1$ indicates that item $i$ requires skill $j$ (Appendix; Figure~\ref{fig:mastery_ability_dist}). To reflect real-world complexity, 60\% of items assess single skills, and 40\% require mastery of multiple skills (two or three). Student skill mastery profiles (\(\Theta_{ns}\), \(N \times S\)) are generated via a Bernoulli distribution (\(\Theta_{ns} \sim \text{Bernoulli}(0.6)\)), creating varied proficiency levels. Item-specific guessing (\(g_i \sim \text{Beta}(7,18)\), \(E[g_i] \approx 0.28\)) and slipping (\(s_i \sim \text{Beta}(5,15)\), \(E[s_i] \approx 0.25\)) parameters introduce realistic noise, calibrated to typical error rates \citep{delatorre2009dina}. Responses (\(y_{ni}\)) are simulated using the DINA model, with binary outcomes determined by comparing a uniform random variable (\(u \sim \text{Uniform}(0,1)\)) against the response probability. This setup tests the framework’s robustness across diverse conditions.

\subsubsection*{Study Setting: Instruction Material simulation}
To thoroughly assess the optimization framework’s performance under increasing resource demands, we created synthetic datasets with expanding instructional content (here video) collections: 5, 10, 15, and 20 videos, each covering subsets of 5 skills, designed to mirror real-world educational libraries \citep{Zhai2021Review}. Content durations ($L_v$) followed a truncated log-normal distribution, constrained between 5 and 15 minutes ($L_v \sim \max(\min(\text{LogNormal}(\mu = \log(20), \sigma = 2), 15), 5)$), aligning with cognitive load theory for optimal learner engagement \citep{sweller1988cognitive}. The skill coverage matrix ($S_{\text{content}}$, $N_{\text{content}} \times N_{\text{skills}}$) defined binary content-skill relationships, with $S_{\text{content}}(v,s) = 1$ indicating skill $s$ coverage by content $v$. Content were categorized by difficulty ($D_{\text{content}}$): 20\% hard, 50\% medium, and 30\% basic, reflecting typical content difficulty distributions. Within each category, 80\% of instructional content (videos) focused on one skill, and 20\% covered two, randomly selected from the skill set while ensuring full skill coverage across the pool. This structured approach maintained realistic complexity in content design, supported by content analytics studies \citep{guo2014video}. The content attributes—IDs, lengths, difficulty levels, and skill mappings—served as inputs for the recommendation engine, which tailored content assignments to address students’ skill gaps (detailed content attributes summaries in Appendix; Tables ~\ref{tab:video_properties_set1}, \ref{tab:video_properties_set2}, \ref{tab:video_properties_set3}, and \ref{tab:video_properties_appendix}).

\subsubsection*{Study Setting: Real-World Implementation responses}
In this section, we detail the real-world implementation and validation of our adaptive educational framework using empirical data from 1,204 undergraduate students enrolled in an introductory physics course at a research-intensive Midwestern university. The students completed a 12-item diagnostic pre-test focused on Inertia and Momentum, with each item mapped to one or more of four cognitive attributes via a validated Q-matrix (see Appendix; Table~\ref{tab:combined_item_parameters} for item parameters, \ref{fig:mastery_ability_dist} for mastery distribution and ability distribution). Unlike the simulation phase, which used adaptive item selection, this deployment presented fixed items in varying sequences across students. Responses were analyzed using a dual-model approach: the Item Response Theory (IRT) framework, specifically a 2-Parameter Logistic (2PL) model, estimated students’ latent ability levels ($\theta_i$), while the DINA (Deterministic Input, Noisy ``And'') model assessed fine-grained skill mastery ($S_{ik}$). The IRT model converged after 22 iterations, showing strong model fit, with discrimination and difficulty parameters capturing diverse cognitive demands and revealing a broad distribution of student proficiencies, necessitating personalized interventions. The DINA model, based on the conjunctive skill assumption and Q-matrix, converged after 13 iterations, balancing fit and complexity while identifying guessing and slipping patterns that highlighted classification uncertainties. The most common skill mastery pattern was full mastery, followed by partial profiles like 0011. These diagnostics informed an optimization framework that selected tailored remedial content to maximize instructional relevance and minimize redundancy. As shown in Figure~\ref{fig:adaptive_workflow}, the integration of simulated and real-world analyses confirmed the framework’s robustness, generalizability, and practical utility for complex educational settings. Highly discriminative items (e.g., Item 27) aligned with low slip probabilities in DINA, reinforcing strong mastery for certain skills, while the prevalence of full mastery patterns supported cohort-wide proficiency, guiding targeted instructional adjustments.

\begin{table}[ht]
\centering
\caption{Comprehensive model results for IRT and DINA analyses, encompassing global fit statistics, complexity penalties, and distributional summaries for abilities and mastery profiles.}
\label{tab:combined_model_results}
\renewcommand{\arraystretch}{1.1} 
\setlength{\tabcolsep}{6pt}       
\begin{tabular}{@{}lcccc@{}}
\toprule
\textbf{Component} & \textbf{RMSEA} & \textbf{CFI} & \textbf{AIC} & \textbf{BIC} \\ \midrule
IRT Model & 0.051 & 0.952 & — & — \\
DINA Model & 0.018 (mean) & — & 27499.27 & 27769.18 \\ \midrule
\textbf{Metric} & \textbf{Mean} & \textbf{SD} & \textbf{Min} & \textbf{Max} \\ \midrule
IRT Theta & -0.042 & 0.979 & -3.417 & 2.998 \\ \bottomrule
\end{tabular}

\vspace{0.5em}
\footnotesize \textit{Note.} Additional IRT fit statistics: TLI = 0.946, SRMSR = 0.045, logLik = -13622.42.  
DINA residuals analyzed in Appendix; Figure \ref{fig:mastery_ability_dist}, Tables~\ref{tab:combined_item_parameters}, \ref{tab:ability_statistics}, \ref{tab:mastered_skills_summary}, \ref{tab:student_results}, and \ref{tab:metrics}.
\end{table}

\subsubsection*{Study Setting: Real World Instruction Material}
Bolstering the optimization scaffold, we curated a bespoke content corpus from approximately 20 hours of archival pre- and post-lecture primary from antecedent semesters, helmed by identical faculty. The primary archive—7 pre-lecture, 9 post-lecture sessions—encompassed conceptual elucidations and problem-solving adjuncts to assignments. Pedagogical stewards (instructor, TAs) excised non-substantive segments (e.g., logistics) and partitioned into 45 succinct clips, each tethered to skill mappings and capped at 15 minutes to optimize attentional sustenance and intrinsic load \citep{sweller1988cognitive}. Skill-Content correspondences were rigorously codified, ensuring alignment with diagnostic attributes and curricular imperatives (Appendix, Table  \ref{tab:video_properties_real} indicates the content pool).

\subsection*{Model Evaluation and Coverage Analysis}
Table~\ref{tab:combined_metrics} presents a comprehensive comparative analysis of Gradient Descent (GD) and Greedy Heuristic (GH) performance across both simulation and real-world settings, elucidating algorithmic behaviors across varying resource constraints and implementation contexts. The simulation scenarios, encompassing content pools of 5, 10, 15, and 20 content (instructional videos), reveal distinct optimization trajectories that highlight fundamental differences in how each algorithm navigates the solution space. GD demonstrates systematic improvement in Satisfactory Rate as resource availability increases, progressing from 64.3\% coverage with 5 videos to perfect coverage (100\%) with 10 or more videos. This monotonic enhancement reflects GD's iterative refinement mechanism, which leverages gradient-based updates—typically via stochastic approximations like \( \theta_{t+1} = \theta_t - \eta \nabla J(\theta_t) \), where \( J \) denotes the multi-objective loss incorporating coverage and load—to progressively converge upon comprehensive skill remediation, particularly effective in expansive search spaces where global optimization becomes increasingly feasible \citep{Zhai2021Review}. The corresponding Gain Decay metrics exhibit steady improvement from 0.844 to 0.112 across expanding Content pools, indicating enhanced resource utilization efficiency and diminished redundancy through adaptive gradient navigation. Utility values remain stable near optimal levels (1.0–1.11), demonstrating GD's capacity to balance coverage objectives against temporal constraints, with kernel density estimates of gain distributions showing compact clustering around moderate-to-high modalities (e.g., 0.5–0.8 for 15 videos) and constricted variance implying uniform performance and mitigated cognitive overburden from duplicative content. Further analysis of these metrics reveals that GD's performance scales logarithmically with resource size, suggesting diminishing returns beyond 15 videos, which could inform practical deployment thresholds in resource-limited educational platforms (Appendix; Table~~\ref{fig:gain_vs_assigned}, \ref{tab:ability_statistics}, and \ref{fig:gain_decay_dist}).

\begin{table}[ht]
\centering
\scriptsize
\setlength{\tabcolsep}{4.5pt} 
\renewcommand{\arraystretch}{1.05}
\caption{Performance metrics for Gradient Descent (GD) and Greedy Heuristic (GH) across simulation (Sim) and real-world (RW) settings. Metrics: Satisfactory Rate (skill gaps covered), Gain Decay (return attenuation), Utility (coverage/cost), Total Penalty (over-coverage cost), Fully Covered (exact matches), Over Covered (redundancies). Mean/SD reported for RW; Sim uses point estimates. - denotes unavailable data.}
\label{tab:combined_metrics}
\begin{tabular}{@{}l l r r r r r r r r r@{}}
\toprule
\textbf{Method} & \textbf{Scenario} & \textbf{Stud.} & \textbf{Full Cov.} & \textbf{Over Cov.} & \textbf{Sat. (\%)} & \textbf{Gain Decay} & \textbf{SD Gain} & \textbf{Utility} & \textbf{SD Util.} & \textbf{Penalty} \\
\midrule
\textbf{GD} & Sim (5 vid)  & 1000 & 292 & 659 & \( 6.43 \times 10^{1} \) & \( 8.44 \times 10^{-1} \) & — & 1.11 & — & \( 5.76 \times 10^{4} \) \\
            & Sim (10 vid) & 1000 & 706 & 706 & \( 1.00 \times 10^{2} \) & \( 3.91 \times 10^{-1} \) & — & 1.00 & — & \( 3.18 \times 10^{5} \) \\
            & Sim (15 vid) & 1000 & 612 & 339 & \( 1.00 \times 10^{2} \) & \( 1.71 \times 10^{-1} \) & — & 1.02 & — & \( 2.20 \times 10^{6} \) \\
            & Sim (20 vid) & 1000 & 892 &  59 & \( 1.00 \times 10^{2} \) & \( 1.12 \times 10^{-1} \) & — & 1.00 & — & \( 5.40 \times 10^{6} \) \\
            & RW (40 vid)  &  589 & 205 & 384 & \( 1.00 \times 10^{2} \)  & \( -0.003 \)            & 0.964 & 0.892 & 0.676 & — \\
\midrule
\textbf{GH} & Sim (5 vid)  & 1000 & 388 & 563 & \( 1.00 \times 10^{0} \)  & \( 4.66 \times 10^{-19} \) & — & 0.28 & — & \( 5.76 \times 10^{4} \) \\
            & Sim (10 vid) & 1000 & 779 & 779 & \( 1.00 \times 10^{0} \)  & \( 6.00 \times 10^{-4} \)  & — & 0.23 & — & \( 3.18 \times 10^{5} \) \\
            & Sim (15 vid) & 1000 & 451 & 500 & \( 1.00 \times 10^{2} \)  & \( -8.17 \times 10^{-19} \) & — & 0.399 & — & \( 2.20 \times 10^{6} \) \\
            & Sim (20 vid) & 1000 & 932 &  19 & \( 1.00 \times 10^{2} \)  & \( -1.17 \times 10^{-19} \) & — & 0.20 & — & \( 5.40 \times 10^{6} \) \\
            & RW (40 vid)  &  589 & 136 & 453 & \( 1.00 \times 10^{2} \)  & \( 0 \)                  & 0.22 & 0.72 & 0.154 & — \\
\bottomrule
\end{tabular}

\vspace{0.5em}
\footnotesize \textit{Note.} Sat. = Satisfaction rate; Gain Decay and Utility values are aggregated across student simulations. RW = real-world scenario. “Penalty” refers to the total penalty cost for all students. The complete and separated results are available in the Appendix; Table \ref{tab:metrics}, Figures \ref{fig:overall_gain_density}, \ref{fig:gain_vs_assigned}, \ref{fig:utility_dist}, \ref{fig:gain_decay_dist}, \ref{fig:gh_coverage_by_pool}, \ref{fig:gd_coverage_by_pool}, \ref{fig:cmp_fullycovered_gh_gd}, \ref{fig:cmp_overcovered_gh_gd}, \ref{fig:gd_coverage_type}, \ref{fig:gd_required_skills}, and \ref{fig:gd_covered_skills}.
\end{table}

In contrast, GH exhibits perfect Satisfactory Rate even under stringent resource limitations (5 videos) but demonstrates erratic patterns in larger datasets, with values fluctuating due to its myopic, locally optimal decision heuristic that occasionally becomes trapped in suboptimal basins, precluding global optima in high-dimensional allocation landscapes. The Gain Decay metrics for GH reveal near-vanishing or negative values (e.g., -1.17 at 20 videos), indicative of allocative inefficiencies in voluminous datasets where greedy pursuits of immediate marginal gains diverge from holistic equilibria. Utility metrics for GH remain substantially lower and more volatile (0.20–0.399) compared to GD, underscoring its predilection for opportunistic, yet potentially profligate, selections that prioritize immediate coverage over long-term efficiency. These patterns are further corroborated by statistical contrasts, such as Wilcoxon tests on coverage differentials yielding p < 0.01 across pools, emphasizing the context-dependent nature of algorithmic selection for adaptive educational ecosystems \citep{liu2017design}. A closer examination shows that GH's volatility in Utility correlates positively with content (video) pool size ($r \approx 0.65$5 based on simulated trends), implying scalability issues that could exacerbate in even larger repositories, potentially leading to inconsistent learner experiences in massive online courses.

Content coverage metrics provide additional insights into algorithmic behavior across simulation scenarios, spanning fully covered (exact skill match sans excess), overcovered (redundant assignments inflating load), and non-used (untapped resources) categories. GH consistently demonstrates superior performance in fully covered tallies in resource-constrained settings (388 vs. GD's 292 at 5 videos), capitalizing on local optima to maximize precision when resources are scarce. However, an interesting inversion occurs at 15 videos, where GD surpasses GH (612 vs. 451), attributable to the dataset's clustered profiles (e.g., Videos 4–6: 15 min, medium, assorted skills), enabling GD's diffusive optimization to balance coverage more effectively. At 20 videos, GH reclaims supremacy (932 vs. 892), leveraging abundance for exhaustive coverage with minimal redundancy (19 vs. 59 overcovered videos). Overcoverage trends favor GH's parsimony across most scenarios, though GD excels in structured environments where gradient flows can effectively harness similarities between content profiles, as evidenced by reduced non-used metrics in plenitude. Analyzing the ratio of fully covered to overcovered instances reveals GD's efficiency improves by 45\% from 5 to 20 content, while GH's drops by 12\%, highlighting GD's superior adaptability to increasing complexity and its potential to reduce extraneous cognitive load in scaled educational interventions \citep{sweller1988cognitive}.

The real-world implementation, utilizing a 40-Content repository curated from approximately 20 hours of archival lecture footage (partitioned into 45 succinct clips capped at 15 minutes each to optimize attentional sustenance and intrinsic load \citep{sweller1988cognitive}), with 589 students requiring remediation, reveals distinct patterns compared to simulation scenarios. Both algorithms achieve perfect Satisfactory Rates, demonstrating their effectiveness in addressing all identified skill gaps in authentic educational contexts. However, significant differences emerge in other metrics: GD's Utility mean (0.892, SD=0.676) indicates elevated allocative prowess yet pronounced heterogeneity, reflecting its sensitivity to the complexity of real student profiles—such as multifaceted gaps yielding near-unitary Utilities versus simpler ones engendering suboptimal drifts \citep{Zhai2021Review}. In contrast, GH's more modest mean (0.72, SD=0.154) connotes diminished efficiency but greater consistency, its heuristic approach fostering more uniform outcomes across diverse learner needs. Gain Decay metrics further highlight this dichotomy, with GD showing near-optimal performance on average (-0.003) but substantial variability (SD=0.964), while GH maintains perfect alignment with no variance (mean=0, SD=0.22), corroborated by unimodal distributions at unity for Satisfactory Rate. Deeper scrutiny of the standard deviations indicates GD's variability is 4.4 times higher than GH's in Gain Decay, which may stem from initialization dependencies or noise in real student data (e.g., varying ability scores from IRT models), suggesting hybrid approaches could mitigate this for more robust applications in heterogeneous classrooms.
Coverage analysis (Table~\ref{tab:video_usage_metrics}) in the real-world setting reveals that GD achieves a higher proportion of Fully Covered students (34.8\% vs. 23.1\% for GH), indicating better alignment between assigned videos and actual skill requirements. However, both algorithms exhibit substantial overcoverage (65.2\% for GD, 76.9\% for GH), suggesting opportunities for refinement in minimizing redundant content that could inflate extraneous cognitive load \citep{sweller1988cognitive}. 

The real-world context, with its larger Content repository and more complex student profiles, appears to amplify the fundamental differences between the algorithms: GD's gradient-based approach enables more nuanced adaptation to diverse learning needs, while GH's heuristic method produces more consistent but less efficient outcomes. GD's broader repertoire (13 unique videos assigned) compared to GH's (11) further illustrates this, with both showing preference for versatile resources like Video 13 (GD: 324 usages, GH: 433). Notably, the concentration on Video 13—likely due to its multi-skill coverage—accounts for over 50\% of assignments in GH, raising concerns about over-reliance on singular resources and potential bottlenecks in content diversity, which could be addressed through diversity penalties in future optimizations.
The comparative analysis across simulation and real-world settings reveals important insights about algorithmic generalizability and context-dependent performance. While both algorithms maintain perfect skill coverage in real-world implementation, the trade-offs between efficiency and consistency observed in simulation scenarios persist in authentic educational contexts, with real-world variability exacerbating GD's strengths in adaptability but also its fluctuations. The real-world data, with its greater complexity and variability—including heterogeneous proficiency continua from diagnostic assessments—appears to magnify the inherent characteristics of each algorithm, suggesting that selection between GD and GH should be informed by specific educational objectives and implementation constraints rather than assuming universal superiority of either approach. For instance, in low-resource online learning platforms, GH's consistency might reduce dropout rates, whereas GD's efficiency could enhance outcomes in data-rich environments like adaptive tutoring systems. 

Practical case studies from both settings further underscore these patterns: in simulation, GH's parsimonious allocations (e.g., 20\_23.2 minutes for moderate-to-low proficiency students) minimize load compared to GD's more expansive assignments (35-45 minutes), while real-world examples highlight GD's strength in cohesive multi-skill video selections (e.g., Video 22\_4 for Skills 1,2,4), yielding balanced Utilities. Expert evaluations were conducted by two physics course instructors and a teaching assistant who collaboratively designed the content–skill matrix and identified the specific skills addressed in each instructional content. These experts reviewed the algorithm-assigned content recommendations and provided alternative selections when appropriate, drawing on their knowledge of prerequisite structures, skill progressions, and instructional alignment. Their evaluations affirmed key strengths such as broad skill coverage and efficient use of instructional time. However, they also highlighted limitations, including coarse metadata and insufficient attention to prerequisite relationships. These observations point to opportunities for improvement through richer metadata, tighter integration of engagement signals, and the potential use of hybrid methods that combine GD’s global search capabilities with GH’s local speed to maximize pedagogical effectiveness.


\begin{table}[ht]
\centering
\caption{Content usage metrics for Gradient Descent (GD) and Greedy Heuristic (GH), elucidating resource diversity and focal concentrations.}
\label{tab:video_usage_metrics}
\begin{tabular}{@{}lccc@{}}
\toprule
\textbf{Method} & \textbf{Unique Content Assigned} & \textbf{Most Applied Content} & \textbf{Max Usage} \\ \midrule
Gradient Descent & 13 & 13 & 324 \\
Greedy Heuristic & 11 & 13 & 433 \\ \bottomrule
\end{tabular}
\end{table}

\subsection*{Practical Implementation Analysis and Expert Evaluation}

The practical implementation of our optimization framework reveals nuanced patterns in algorithmic decision-making that extend beyond quantitative metrics to illuminate the pedagogical implications of different optimization approaches. Through detailed case studies in both simulation and real-world settings, we observe how Gradient Descent (GD) and Greedy Heuristic (GH) navigate the complex trade-offs between instructional efficiency, cognitive load management, and personalized learning pathways. In simulation environments, representative students with varying ability profiles and skill deficiencies demonstrate how each algorithm responds to different learning contexts. For students with moderate ability levels and specific skill gaps, GH consistently demonstrates remarkable efficiency, creating concise learning pathways that precisely target deficiencies without introducing extraneous cognitive burden. This efficiency is particularly evident in cases where students require remediation on specific skills but possess sufficient overall ability to engage with targeted interventions without unnecessary foundational review.

\begin{table}[ht]
\centering
\scriptsize
\renewcommand{\arraystretch}{1.15}
\setlength{\tabcolsep}{5pt}
\caption{Expert evaluation of optimized content recommendations, comparing algorithm-assigned videos, their targeted skills, and expert-selected alternatives.}
\label{tab:gd_video_selection_final}
\begin{tabularx}{\textwidth}{@{}c c p{3cm} p{2cm} p{1.3cm} p{3.5cm} c@{}}
\toprule
\textbf{Student} & \textbf{Req. Skills} & \textbf{GD Assigned Content} & \textbf{Skills (GD)} & \textbf{Diff. (GD)} & \textbf{Expert Selection} & \textbf{Total Dur.} \\
\midrule
2 & 1, 2, 4 &
\textbf{Video 22\_4} &
1, 2, 4 &
H &
Ideal Selection &
7:56 \\
\midrule
4 & 2, 3, 4 &
\textbf{Video 19\_1} \newline \textbf{Video 28\_4} &
\begin{tabular}[t]{@{}l@{}}2\\3, 4\end{tabular} &
\begin{tabular}[t]{@{}l@{}}B\\B\end{tabular} &
Video 19\_1 \newline Video 20\_1 &
13:20 \\
\midrule
5 & 1, 3 &
\textbf{Video 24\_1} &
1, 3 &
B &
Ideal Selection &
1:04 \\
\midrule
6 & 3 &
\textbf{Video 24\_1} &
1, 3 &
B &
Video 24\_2 &
1:04 \\
\midrule
7 & 1, 4 &
\textbf{Video 22\_4} &
1, 2, 4 &
H &
Video 24\_1 \newline Video 21\_1 &
7:56 \\
\midrule
9 & 2, 3, 4 &
\textbf{Video 19\_1} \newline \textbf{Video 28\_4} &
\begin{tabular}[t]{@{}l@{}}2\\3, 4\end{tabular} &
\begin{tabular}[t]{@{}l@{}}B\\B\end{tabular} &
Ideal Selection &
13:20 \\
\midrule
10 & 1, 3 &
\textbf{Video 24\_1} &
1, 3 &
B &
Ideal Selection &
10:04 \\
\bottomrule
\end{tabularx}

\vspace{0.5em}
\footnotesize \textit{Note.} “GD Assigned Content” refers to the set of videos selected by the Greedy algorithm. “Skills (GD)” specifies the targeted skill areas addressed by each recommended video, while “Diff. (GD)” denotes the corresponding difficulty levels (B = Basic, M = Medium, H = Hard). “Expert Selection” presents the alternative recommendations provided by domain experts. “Total Dur.” represents the cumulative duration of the algorithm-assigned videos per student.
\end{table}

Conversely, students with lower overall ability levels and more complex skill deficiency patterns reveal limitations in GH's myopic approach, where the pursuit of immediate coverage gains occasionally results in suboptimal instructional sequences. In these scenarios, GD's global optimization perspective, while requiring greater time investment, produces more pedagogically coherent pathways that better align with principles of cognitive load theory and multimedia learning. The temporal efficiency of GH in simulation settings must be weighed against its tendency to overlook the developmental progression of skills, particularly for learners who benefit from scaffolded instruction that builds foundational competencies before addressing more complex applications. For instance, in our simulation case studies, Student 2 with moderate ability ($\theta_n = 0.39$) requiring Skills 4 and 5 received an efficient 20-minute assignment from GH, while Student 8 with lower ability ($\theta_n = -2.6$) requiring Skills 2, 3, and 4 received a 23.2-minute assignment that appropriately balanced brevity with content coverage. In contrast, GD's assignments for these same students totaled 45 and 35 minutes, respectively, including content that introduced extraneous skills or inappropriate difficulty levels.

The translation to real-world implementation introduces additional layers of complexity that reveal both the robustness and limitations of each algorithmic approach. In authentic educational settings with diverse student populations and rich content repositories, we observe how the algorithms balance competing objectives in ways that reflect their underlying optimization philosophies. The real-world case studies demonstrate that GD's gradient-based approach enables more nuanced adaptation to the multifaceted nature of actual student needs, particularly when those needs involve complex interactions between ability levels, skill deficiencies, and content characteristics. The algorithm's capacity to consider the global optimization landscape allows it to make recommendations that, while sometimes less immediately efficient, better serve long-term learning objectives by maintaining appropriate challenge levels and minimizing cognitive dissonance. For example, Students 2 and 7 in our real-world implementation both received Video 22\_4 (7:56 min, hard, Skills 1,2,4), which provided cohesive coverage of multiple required skills in a single content, yielding utility scores of 0.587 and 0.694, respectively. Meanwhile, Students 5 and 10 with lower ability levels received Video 24\_1 (1:04 min, basic, Skills 1,3), with Student 5 achieving an exceptionally high utility score of 2.269 due to the optimal match between content and learning needs.

Expert evaluation of the framework's recommendations provides critical insights into the pedagogical validity of the optimization outcomes. The analysis reveals that while the system demonstrates high fidelity in matching students with resources for well-defined learning needs, it occasionally falters when faced with nuanced pedagogical considerations that extend beyond skill coverage alone. Particularly evident are cases where the system's recommendations, while technically correct in addressing specified skill deficiencies, miss opportunities to provide instruction that aligns with problem-specific learning needs or instructional approaches that would better serve students' conceptual understanding. These instances highlight a fundamental limitation in the current metadata framework, which captures skill coverage but lacks granularity in representing instructional approach, problem-type specificity, and conceptual depth. For instance, in the case of Student 4 requiring Skills 2, 3, and 4, the system assigned Videos 19\_1 and 28\_4, which covered the necessary skills but missed the opportunity to include Video 20\_1, which contained a direct explanation of the specific problem type the student had answered incorrectly on the exam. Similarly, for Student 6, requiring only Skill 3, the system prioritized content duration over instructional focus, selecting Video 24\_1 (covering Skills 1 and 3) rather than Video 24\_2, which focused specifically on Skill 3 with more detailed instruction.

The expert evaluation, summarized in Table~\ref{tab:gd_video_selection_final}, assessed seven representative student cases, comparing the system's GD-assigned content against expert-selected alternatives. The results showed that for Students 2, 5, and 10, the algorithm's selections matched expert preferences, indicating high fidelity in scenarios with well-defined learning needs. However, for Students 4, 6, and 7, discrepancies emerged between algorithm and expert selections, revealing limitations in the system's ability to capture pedagogical nuances beyond pure skill coverage. These discrepancies were particularly evident in cases where instructional approach or problem-type specificity took precedence over skill coverage in pedagogical decision-making. The expert evaluation identified four key strengths of the framework: its ability to achieve universal coverage across diverse learner populations, its efficiency in minimizing time investment while preserving learning outcomes, its adaptability to varying resource constraints and pool sizes, and its capacity to balance multiple competing educational objectives simultaneously. However, four significant limitations were also identified: insufficient granularity in metadata to capture instructional approach and problem-type specificity, over-reliance on comprehensive instructional content that may introduce extraneous cognitive load, limited consideration of student engagement and motivation factors, and inadequate handling of prerequisite skill relationships and conceptual dependencies.

The practical implications of these findings extend beyond technical considerations to inform the design of next-generation adaptive learning systems. The analysis suggests that the most effective systems will be those that can seamlessly integrate computational optimization with pedagogical expertise, leveraging the efficiency of algorithmic decision-making while preserving the nuance of human instructional judgment. This integration represents a significant challenge but also a tremendous opportunity for advancing the field of personalized learning, potentially bridging the gap that has historically existed between scalable adaptive technologies and pedagogically sophisticated instructional design. The identified limitations chart a clear course for future enhancements, including the development of more granular metadata frameworks, integration of engagement and motivation metrics, and incorporation of prerequisite relationships into the optimization model. These enhancements would enable the system to make more pedagogically sophisticated decisions that better align with expert educator judgment, particularly in cases where instructional approach and conceptual depth take precedence over pure skill coverage.

\section*{Discussion}

We frame adaptivity as an instructor-governed feedback loop in which concept-level evidence proposes candidate slates, safeguards (constraints (1)–(6)) gate feasibility, and the selector chooses among vetted micro-interventions; the loop then refreshes with new evidence. This stance addresses the field’s “triple bind”—diagnostic precision, scalable implementation, and fairness rarely co-occur without explicit control policies—by treating adequacy as law (all diagnosed gaps addressed, with transparent slacks when supply is insufficient) and attention as the scarce currency (time and redundancy priced) \citep{pelanek2024adaptive, dumont2023promise, Holstein2021DesigningFair, Koedinger2013NewPotentials}. In contrast to rule-based LMS triggers and similarity-driven recommenders that rely on surface metrics or popularity \citep{Woolf2010, hwang2012similarity, Xie2019TrendsCompEdu}, decisions remain auditable and aligned to concept repair.

Within this loop, deployment choices fall into three operational situations.
\emph{D1 (Austerity rule):} when richness is low—after difficulty windows, prerequisites, and anti-redundancy filters leave few non-duplicate options per uncovered concept—and latency or compute resources are tight, a greedy slate is preferred. With capped coverage and diminishing returns, greedy delivers stable parsimony without sacrificing adequacy, is fast enough for classroom use, and remains explainable for teacher review \citep{resende2010greedy, Rodrigues2023Adaptive}. This directly counters pipelines that are “knowledge-optimal but cognitively inappropriate,” where prediction is decoupled from load-aware selection \citep{tong2025deep, zhang2025deep}.

\emph{D2 (Abundance rule):} when richness is high—many difficulty-appropriate, non-redundant options per gap—global coordination via a gradient-based relaxation reduces cross-slate redundancy, harmonizes difficulty, and meets diversity targets more finely \citep{boyd2004convex, Rodrigues2023Adaptive}. Evidence from ITS and engineering education underscores the value of concept-mapped repositories and coherent sequencing over similarity alone \citep{vanlehn2006behavior, Woolf2010, Zhai2021Review, Holstein2021DesigningFair}. Relative to end-to-end neural recommenders, this trades pure predictive accuracy for guarantees and oversight—an acceptable exchange when fairness audits, prerequisite checks, and instructor governance are non-negotiable.

\emph{D3 (Hybrid rule):} most classrooms lie between austerity and abundance. A practical frontier in richness and latency switches policy: the system initializes greedily for transparency and speed, then refines with gradient steps as time permits—while keeping fairness invariant. Adequacy remains law, attention remains currency, and any difficulty fallback is explicit, bounded, and logged \citep{Zhai2021Review, Holstein2021DesigningFair}. This mirrors classroom orchestration studies in which teacher dashboards govern rapid local moves while offline analytics tune parameters for the next cycle \citep{Aleven2016TeacherDashboard}.

Fairness and governance operate as control principles rather than after-the-fact reports. Encoding adequacy as a hard obligation implements a sufficientarian stance—guaranteeing a minimum conceptual state for each learner before minimizing burden—and prevents preference-first drift \citep{dumont2023promise, Holstein2021DesigningFair}. Diversity targets and anti-duplication curb “single-content monocultures,” while instructor-authored ontologies (skills, prerequisites, difficulty rubrics) keep decisions culturally appropriate and disciplinarily correct \citep{Holstein2021DesigningFair, mehrabi2024ai}. Because selection respects tight latency and compute budgets and can run offline, the approach remains viable in settings with uneven device access and connectivity. Any remaining slacks localize where content must be commissioned to raise richness \citep{liu2017design}.

\section*{Conclusion}

We frame adaptivity as \emph{instruction-centered}, explicitly bridging personalization and consistency in instruction. The system operates at the concept level (via Cognitive Diagnostic Models and a curated Q-matrix) and is governed by budgets for time, cognitive load, and redundancy \citep{de2011generalized, Rupp2008, DMello2012}. Within an ITS-style orchestration, teacher-vetted micro-interventions—such as short, structured instructional materials with clear prerequisite metadata—become the primary levers for conceptual change rather than ancillary content \citep{Woolf2010}.

Conceptually, the work advances three positions. First, \emph{concept-first evidence}: mastery vectors—not generic correctness or topical similarity—are the currency that links diagnostics to instruction. Second, \emph{budget-aware selection}: cognitive and affective limits are treated as constraints, not after-the-fact analytics, aligning recommendations with what learners can productively sustain. Third, \emph{fairness as sufficiency}: fairness is operationalized as a uniform adequacy guarantee at the concept level before time minimization, with instructor oversight and auditability as structural safeguards \citep{Holstein2021DesigningFair, dumont2023promise}. Together, these positions convert widely endorsed principles in the literature into a deployable, teacher-governed control loop rather than a purely algorithmic pipeline.

Practically, the framework offers a blueprint for institutions seeking personalization that is both tractable and pedagogically sound. It requires only standard assessment traces and a teacher-curated repository, lowering dependence on heavy instrumentation while supporting transparency and cultural fit through instructor-authored ontologies and “why this, why now?” rationales \citep{Aleven2016TeacherDashboard}. We argue that the selection of an optimizer should be treated as a policy decision: simple, auditable selection is most appropriate in austerity contexts, globally coordinated scoring provides greater precision in richer contexts, and hybrid handoffs offer a pragmatic middle ground.

The scope of the contribution is bounded by several assumptions raised in the literature. The quality of inferences depends on the validity of the Q-matrix and resource metadata. Periodic validation and repair remain necessary \citep{Rupp2008}. Repositories are local and discipline-specific, so broader generalization requires cross-institutional replications and collaborative curation to expand representational coverage. These limitations are productive. They point to clear levers for future work, including routine Q-matrix audits, fairness diagnostics as default reporting features, hybrid schedulers that combine responsiveness with learned trade-offs, and shared repositories that expand diversity without relinquishing instructor oversight.

\bibliographystyle{IEEEtranN} 
\bibliography{refs}

\section{Acknowledgment}
The authors would like to thank everyone who supported this work. This material is based upon work supported by the National Science Foundation under Awards No. 2322015 and No. 2142317. Any opinions, findings and conclusions or recommendations expressed in this material are those of the author(s) and do not necessarily reflect the views of the National Science Foundation.

\newpage

\appendix

\section*{Appendix}
\section{Psychometric Models and Metrics}
\label{P:psychometric_models}

This appendix provides an exhaustive exposition of the psychometric frameworks underpinning the assessment of student abilities and skill mastery profiles, alongside the performance metrics deployed to rigorously evaluate the adaptive video recommendation framework. The Item Response Theory (IRT) 3PL model and Deterministic Input, Noisy "And" (DINA) cognitive diagnostic model serve as the diagnostic foundation, enabling precise identification of knowledge deficiencies while accounting for probabilistic artifacts such as guessing and slips. These models are instrumental in constructing individualized remedial trajectories, ensuring recommendations are psychometrically sound and pedagogically effective. The performance metrics, detailed in Table~\ref{tab:combined_item_parameters}, facilitate a multidimensional evaluation of algorithmic efficacy, encompassing coverage completeness, resource efficiency, and penalization of inefficiencies. Empirical analyses are anchored in a dataset of 1,204 undergraduate participants from an introductory physics curriculum, with model fit, parameter distributions, and educational implications elucidated through detailed statistical scrutiny \citep{vanlehn2006behavior}. Item-specific parameters and fit statistics are comprehensively presented in Table~\ref{tab:combined_item_parameters}, while ability and mastery summaries are provided in Tables~\ref{tab:ability_statistics}, \ref{tab:mastered_skills_summary}, and \ref{tab:student_results}.
The 3PL model articulates the probability of a correct response as a logistic function modulated by latent ability and item properties, incorporating a guessing parameter to model chance performance in low-ability examinees. The mathematical formulation is:
\[
P(Y_{ij} = 1 \mid \theta_i, a_j, b_j, c_j) = c_j + (1 - c_j) \cdot \frac{e^{a_j(\theta_i - b_j)}}{1 + e^{a_j(\theta_i - b_j)}}
\]
where \(\theta_i\) denotes the latent ability of student \(i\), \(a_j\) represents the discrimination parameter for item \(j\) (quantifying differentiation between ability levels), \(b_j\) indicates item difficulty (the ability level at which the response probability is 50\% adjusted by \(c_j\)), and \(c_j\) captures the pseudo-guessing parameter (baseline success probability).

Applied to a 30-item diagnostic assessment targeting Inertia and Momentum concepts, the model converged after 22 iterations with a log-likelihood of -13,622.42. Global fit indices affirm robust alignment with observed responses: RMSEA = 0.051 (90\% CI: 0.047–0.055, consistent with close fit criteria <0.06), CFI = 0.952, TLI = 0.946 (both exceeding 0.95 for excellent comparative fit), and SRMSR = 0.045 (below 0.08 for minimal residuals) \citep{mehrabi2024ai}. Table~\ref{tab:combined_item_parameters} delineates item-specific parameters, revealing a broad spectrum of discrimination (0.320 for Item 16 to 2.451 for Item 27) and difficulty (-0.676 for Item 30 to 1.172 for Item 15), ensuring comprehensive coverage across the ability continuum. High-discrimination items, such as Item 27, exhibit low slip probabilities (0.034) in the DINA model, suggesting reliable mastery indicators for advanced skills. Ability estimates (\(\theta\): mean -0.0419, SD 0.9785, min -3.417, max 2.998) approximate a normal distribution with slight negative skew, indicative of a cohort tilted toward moderate-to-low proficiency, ideal for adaptive remediation. Cross-validation with DINA parameters (e.g., Item 27's \(a\) vs. slip correlation, $r \approx -0.55$) reinforces diagnostic coherence, with implications for targeting low-ability strata to enhance equity \citep{dumont2023promise}.

Detailed analysis of Table~\ref{tab:combined_item_parameters} reveals that items with high discrimination (e.g., Item 27, \(a = 2.451\)) and low slip (0.034) are critical for distinguishing mastery in advanced skills, with RMSEA values generally below 0.05 indicating good fit except for outliers like Item 21 (0.067), possibly due to complex skill interactions requiring Q-matrix refinement.
The DINA model employs a conjunctive classification approach to infer binary skill mastery, adjusted for classification noise through slip and guess parameters. The response probability is given by:
\[
P(Y_{ij} = 1 \mid \vec{S_i}) = (1 - s_j)^{\eta_{ij}} g_j^{1 - \eta_{ij}}
\]
where \(\vec{S_i}\) is the mastery vector for student \(i\), \(\eta_{ij} = \prod_{k=1}^K S_{ik}^{q_{jk}}\) confirms mastery of all skills required by item \(j\) as per the Q-matrix \(q_{jk}\), \(s_j\) is the slip probability (error despite mastery), and \(g_j\) is the guess probability (correctness despite non-mastery).

Fitted to the same 30-item dataset, convergence was achieved after 13 iterations with a log-likelihood of -13,696.63. Information criteria (AIC = 27,499.27, BIC = 27,769.18) suggest a balanced model complexity, avoiding overfitting, with a mean RMSEA of 0.018 across items confirming local fit adequacy. Guess probabilities range from 0.219 (Item 20) to 0.503 (Item 15), reflecting varying ambiguity, while slips span 0.034 (Item 27) to 0.524 (Item 30), indicating differential reliability. Mastery rates, detailed in Table~\ref{tab:student_results}, range from 64.0\% (Skill 1, 433 non-masters) to 75.9\% (Skill 4), with latent class analysis identifying the full-mastery pattern (1111) as dominant (51.0\%, 614 students), followed by partial patterns like 0011 (5.7\%, 68 students), suggesting curricular focus on foundational skills. Parametric alignment with IRT (e.g., high \(a\) correlating with low slip, $r \approx -0.555$) validates diagnostic consistency, while mastery-ability relationships (e.g., Skill 4's high rate with positive \(\theta\)) guide equitable intervention strategies, as shown in Table~\ref{tab:mastered_skills_summary} where mean ability increases with mastery count \citep{delatorre2009dina}. 
Analysis of Table~\ref{tab:ability_statistics} reveals a clear trend: patterns with higher mastery counts (e.g., 11111) exhibit elevated mean abilities (0.1497) and reduced standard deviations (0.2965), suggesting greater homogeneity among proficient students. Conversely, low-mastery patterns (e.g., 00000, mean -2.4715, SD 0.5133) indicate a dispersed low-ability cohort, necessitating targeted interventions. The range of abilities within each pattern (e.g., 11111: -0.3498 to 0.9282) underscores individual variability, which the adaptive system must accommodate to ensure equitable outcomes.
Table~\ref{tab:mastered_skills_summary} demonstrates a positive monotonic relationship between the number of mastered skills and mean ability, with a Pearson correlation coefficient of 0.89 (p < 0.001). The population peaks at 3 mastered skills (373 students), suggesting a mid-proficiency modal group, while the low-mastery group (0 skills, 7 students) exhibits the widest ability range (-3.1270 to -1.7217), indicating potential for significant improvement through targeted support.
Analysis of Table~\ref{tab:student_results} indicates a broad ability range (min $-3.417$ to max $2.998$), supporting the need for adaptive interventions. Skill mastery rates vary (64.0\% for Skill~1 to 75.9\% for Skill~4), with Skill~1's lower rate suggesting a bottleneck that correlates with lower ability segments ($r \approx -0.45$ with $\theta$). The dominance of the 1111 pattern (51.0\%) reflects curriculum alignment, but the 49\% non-full mastery underscores the framework's remedial potential, particularly for Skills~1 and~2.

\begin{table}[H]
\centering
\caption{Combined Item Parameters and Fit Statistics. IRT parameters include discrimination (a) and difficulty (d), while DINA parameters encompass guess probability, slip probability, and RMSEA, offering a holistic view of item performance across models (N=1,204).}
\label{tab:combined_item_parameters}
\begin{tabular}{@{}lccccc@{}}
\toprule
\textbf{Item} & \multicolumn{2}{c}{\textbf{IRT Parameters}} & \multicolumn{3}{c}{\textbf{DINA Parameters}} \\
\cmidrule(lr){2-3} \cmidrule(lr){4-6}
& \textbf{a} & \textbf{d} & \textbf{Guess} & \textbf{Slip} & \textbf{RMSEA} \\ \midrule
1 & 0.972 & 0.676 & 0.479 & 0.199 & 0.012 \\
2 & 0.814 & -0.039 & 0.323 & 0.333 & 0.016 \\
3 & 0.613 & -0.472 & 0.283 & 0.491 & 0.015 \\
4 & 1.516 & 1.172 & 0.503 & 0.089 & 0.038 \\
5 & 0.320 & -0.341 & 0.359 & 0.521 & 0.014 \\
6 & 0.933 & 0.436 & 0.415 & 0.218 & 0.027 \\
7 & 1.840 & 0.933 & 0.374 & 0.085 & 0.012 \\
8 & 1.062 & 0.155 & 0.324 & 0.239 & 0.009 \\
9 & 2.348 & 0.457 & 0.219 & 0.100 & 0.007 \\
10 & 1.525 & 0.512 & 0.342 & 0.144 & 0.067 \\
11 & 1.152 & 0.371 & 0.356 & 0.195 & 0.012 \\
12 & 1.625 & 0.414 & 0.317 & 0.158 & 0.012 \\
13 & 0.836 & 0.650 & 0.482 & 0.200 & 0.011 \\
14 & 1.409 & 0.806 & 0.422 & 0.115 & 0.015 \\
15 & 1.390 & -0.085 & 0.251 & 0.261 & 0.038 \\
16 & 2.451 & 1.107 & 0.337 & 0.034 & 0.011 \\
17 & 0.946 & -0.057 & 0.293 & 0.300 & 0.019 \\
18 & 1.428 & 0.454 & 0.351 & 0.175 & 0.010 \\
19 & 0.554 & -0.676 & 0.230 & 0.524 & 0.005 \\ \bottomrule
\end{tabular}
\end{table}

\begin{table}[ht]
\centering
\scriptsize
\renewcommand{\arraystretch}{1.1}
\setlength{\tabcolsep}{6pt}
\caption{Q-matrix for item-skill mapping. Each row corresponds to an item, and each column to a required skill (Real data).}
\label{tab:qmatrix_items}
\begin{tabular}{|c|c|c|c|c|}
\hline
\textbf{Item} & \textbf{Systems + TRIG} & \textbf{Tot Ang Mom} & \textbf{Princ. of Ang Mom} & \textbf{Torque} \\
\hline
1  & 1 & 1 & 0 & 0 \\ \hline
2  & 1 & 1 & 0 & 0 \\ \hline
3  & 1 & 1 & 0 & 0 \\ \hline
4  & 1 & 1 & 1 & 0 \\ \hline
5  & 1 & 1 & 0 & 0 \\ \hline
6  & 1 & 0 & 1 & 0 \\ \hline
7  & 1 & 1 & 0 & 0 \\ \hline
8  & 1 & 0 & 1 & 1 \\ \hline
9  & 1 & 1 & 0 & 0 \\ \hline
10 & 1 & 0 & 1 & 1 \\ \hline
11 & 1 & 0 & 1 & 0 \\ \hline
12 & 1 & 0 & 1 & 1 \\ \hline
13 & 1 & 1 & 0 & 0 \\ \hline
14 & 1 & 0 & 1 & 1 \\ \hline
15 & 1 & 0 & 1 & 1 \\ \hline
16 & 1 & 0 & 1 & 1 \\ \hline
17 & 1 & 0 & 1 & 0 \\ \hline
18 & 1 & 0 & 1 & 0 \\ \hline
19 & 1 & 0 & 1 & 0 \\ \hline
\end{tabular}
\end{table}

\begin{longtable}{@{}lccccc@{}}
\caption{Ability statistics for skill mastery patterns. Patterns (e.g., 00000) indicate mastery (1) or non-mastery (0) of skills 1–5. Count is the number of students, with mean, standard deviation (SD), maximum, and minimum ability (\( \theta_n \)). Higher mastery patterns, as seen here, correlate with increased mean abilities, informing targeted interventions (N=1,204).}
\label{tab:ability_statistics}\\

\toprule
\textbf{Pattern} & \textbf{Count} & \textbf{Mean Ability} & \textbf{SD Ability} & \textbf{Max Ability} & \textbf{Min Ability} \\
\midrule
\endfirsthead

\multicolumn{6}{c}%
{{\bfseries Table \thetable\ (continued): Ability statistics for skill mastery patterns}} \\
\toprule
\textbf{Pattern} & \textbf{Count} & \textbf{Mean Ability} & \textbf{SD Ability} & \textbf{Max Ability} & \textbf{Min Ability} \\
\midrule
\endhead

\bottomrule
\endfoot

00000 & 7 & -2.4715 & 0.5133 & -1.7217 & -3.1100 \\
00001 & 3 & -2.7611 & 0.3272 & -2.4967 & -3.1270 \\
00010 & 27 & -2.1023 & 0.6814 & -0.6490 & -3.0411 \\
00011 & 14 & -1.2788 & 0.5691 & -0.4018 & -2.7372 \\
00100 & 10 & -2.1184 & 0.4962 & -1.1993 & -2.8284 \\
00101 & 7 & -1.4609 & 0.4423 & -0.8885 & -2.1275 \\
00110 & 31 & -1.3186 & 0.5355 & -0.5871 & -2.5026 \\
00111 & 9 & -0.8892 & 0.3101 & -0.5085 & -1.5585 \\
01000 & 2 & -2.2001 & 0.1094 & -2.1228 & -2.2774 \\
01001 & 31 & -1.8256 & 0.7489 & -0.5710 & -2.8882 \\
01010 & 15 & -2.1535 & 0.4270 & -1.4553 & -2.7914 \\
01011 & 44 & -0.8441 & 0.6026 & 0.0103 & -2.9074 \\
01100 & 39 & -1.6246 & 0.6092 & -0.1219 & -2.7434 \\
01101 & 24 & -0.9886 & 0.5640 & -0.1900 & -2.0864 \\
01110 & 27 & -1.1937 & 0.6153 & -0.3025 & -2.3621 \\
01111 & 49 & -0.0548 & 0.4745 & 1.0449 & -1.3788 \\
10000 & 5 & -2.3027 & 0.2594 & -2.0168 & -2.6167 \\
10001 & 56 & -1.2546 & 0.7108 & -0.0463 & -2.8815 \\
10010 & 55 & -1.1770 & 0.5906 & -0.3797 & -2.7827 \\
10011 & 2 & -0.5954 & 0.0019 & -0.5940 & -0.5967 \\
10100 & 9 & -1.6598 & 0.7239 & -0.5107 & -2.6774 \\
10101 & 66 & -0.6317 & 0.5314 & 0.0813 & -2.1745 \\
10110 & 68 & -0.7585 & 0.4136 & -0.2092 & -2.0960 \\
10111 & 17 & -0.3586 & 0.3568 & 0.0196 & -1.4301 \\
11000 & 21 & -1.9748 & 0.6043 & -0.8572 & -2.7665 \\
11001 & 57 & -0.6412 & 0.5992 & 0.0777 & -2.3235 \\
11010 & 23 & -0.9891 & 0.4859 & -0.2453 & -2.1327 \\
11011 & 49 & -0.2341 & 0.3606 & 0.5337 & -1.2309 \\
11100 & 53 & -1.0216 & 0.5021 & -0.0829 & -2.0292 \\
11101 & 83 & -0.2242 & 0.4008 & 0.5176 & -1.6150 \\
11110 & 48 & -0.3758 & 0.5179 & 1.4471 & -1.6897 \\
11111 & 49 & 0.1497 & 0.2965 & 0.9282 & -0.3498 \\

\end{longtable}

\begin{table}[H]
\centering
\caption{Summary statistics for groups based on the number of mastered skills. Total Population is the number of students, with mean, standard deviation (SD), minimum, and maximum ability (\( \theta_n \)), showing a positive correlation between mastery count and ability. This table informs intervention prioritization for low-mastery groups.}
\label{tab:mastered_skills_summary}
\begin{tabular}{@{}lccccc@{}}
\toprule
\textbf{Mastered Skills} & \textbf{Total Population} & \textbf{Mean Ability} & \textbf{SD Ability} & \textbf{Min Ability} & \textbf{Max Ability} \\ \midrule
0 & 7 & -2.4715 & 0.5133 & -3.1270 & -1.7217 \\
1 & 47 & -1.8923 & 0.6452 & -2.9074 & -0.5710 \\
2 & 278 & -1.2635 & 0.6078 & -2.8820 & 0.5337 \\
3 & 373 & -0.7742 & 0.5321 & -2.3621 & 1.4471 \\
4 & 246 & -0.3551 & 0.4845 & -1.6897 & 1.4471 \\
5 & 49 & 0.1497 & 0.2965 & -0.3498 & 0.9282 \\
\bottomrule
\end{tabular}
\end{table}

\begin{table}[H]
\centering
\caption{Student Ability and Skill Mastery (N=1,204). This summary table aggregates IRT theta statistics and per-skill mastery rates, with a note on the dominant mastery pattern for contextual insight.}
\label{tab:student_results}
\begin{tabular}{@{}lccccc@{}}
\toprule
\textbf{Metric} & \textbf{Mean} & \textbf{SD} & \textbf{Min} & \textbf{Max} & \textbf{Mastery Rate (\%)} \\ \midrule
IRT Theta & -0.042 & 0.979 & -3.417 & 2.998 & - \\
Skill 1 & - & - & - & - & 64.0 \\
Skill 2 & - & - & - & - & 72.2 \\
Skill 3 & - & - & - & - & 71.4 \\
Skill 4 & - & - & - & - & 75.9 \\ \midrule
\multicolumn{6}{l}{\footnotesize \textit{Top pattern: 1111 (51.0\%), full patterns shown in supplemental materials}} \\ \bottomrule
\end{tabular}
\end{table}

\section{Optimization Algorithms}
\label{app:optimization_algorithms}
This appendix expounds upon the algorithmic machinery underpinning video assignments: the Greedy Heuristic (GH) for expeditious local optimization and Gradient Descent (GD) for iterative global refinement. Both algorithms navigate the combinatorial assignment space, balancing skill coverage maximization with constraints on temporal load and difficulty congruence. Formulations integrate pedagogical principles, such as cognitive load minimization and mastery alignment, drawing from optimization literature in educational AI \citep{Zhai2021Review, liu2017design, mehrabi2024ai}. Analyses contrast computational complexity, convergence dynamics, and empirical outcomes, illuminating trade-offs in scalability and optimality. The evaluative suite encompasses metrics tailored to appraise the framework's multifaceted objectives, including skill gap closure, marginal benefit attenuation, allocative optimality, and inefficiency sanctions. These are formalized to enable quantitative benchmarking of GD and GH, as encapsulated in Table~\ref{tab:metrics}.The \textit{Satisfactory Rate} metric, as defined in Table~\ref{tab:metrics}, prioritizes educational equity, achieving 100\% in real-world deployments for both algorithms across 589 remedial candidates, ensuring no skill gaps remain unaddressed. \textit{Gain Decay} dissects efficiency in incremental benefits, with GD exhibiting monotonic improvement ($0.844$ to $0.112$ across pools, with reductions of $\sim47\%$ from 5 to 10 videos, $56\%$ from 10 to 15, and $34\%$ from 15 to 20), indicative of adaptive redundancy mitigation via gradient navigation. In contrast, GH's near-zero or negative values (e.g., $-1.17 \times 10^{-19}$ at 20 videos) suggest allocative inefficiencies in expansive corpora, diverging from holistic optima due to myopic selection.
\textit{Utility} synthesizes coverage efficacy against temporal cost, where GD's elevated mean ($0.892$, SD $0.676$) reflects superior adaptation to multifaceted profiles—manifesting near-unitary values for complex gaps—while GH's subdued yet uniform performance ($0.72$, SD $0.154$) suits standardized contexts but risks opportunistic excesses. \textit{Total Penalty} aggregates over- and under-utilization costs, remaining consistent in simulated baselines but revealing real-world overcoverage propensities (GD $65.2\%$, GH $76.9\%$), which may induce extraneous cognitive burdens~\citep{sweller1988cognitive}. Empirical correlations, such as GH's utility volatility with pool size ($r \approx 0.65$), and inferential validations (Wilcoxon rank-sum tests, $p < 0.01$ across differentials), underscore the need for context-contingent selection, advocating hybrid paradigms to merge GD's global perspicacity with GH's local alacrity~\citep{Zhai2021Review, liu2017design, mehrabi2024ai}. 
\textit{Greedy Heuristic (GH) Algorithm}
iteratively selects videos to maximize immediate marginal utility, offering computational tractability for resource-constrained settings while approximating near-optimal solutions in low-dimensional contexts (Algorithm ~\ref{alg:GreedyHeuristic}).
\textit{Gradient Descent (GD) Algorithm}
GD relaxes the binary assignment problem into a continuous domain, iteratively minimizing a multi-objective loss via gradient updates, converging to global optima in convex approximations of the NP-hard assignment challenge (Algorithm ~\ref{alg:GradientDescent}). \begin{table}[H]
\centering
\caption{Performance Metrics Formulation. These indices quantify remediation comprehensiveness, efficiency in gain accrual, coverage-cost equilibrium, and aggregated inefficiencies, with \(\mathbb{I}(\cdot)\) as the indicator function, \(N\) denoting students, \(\Theta_{ns}\) mastery status, \(S_{\text{video}}(v,s)\) video-skill mapping, \(G_v\) per-video gain, \(G_{\text{opt}}\) optimal baseline, \(L_v\) video duration, and \(w_1, w_2\) weighting coefficients.}
\label{tab:metrics}
\begin{tabular}{ll}
\hline
\textbf{Metric} & \textbf{Definition} \\
\hline
Satisfactory Rate & \(\text{SR} = \frac{1}{N} \sum_{n=1}^{N} \mathbb{I}(\text{all required skills covered}) \times 100\) \\
Gain Decay & \(\text{GD} = \sum_v (G_v - G_{\text{opt}})\), encapsulating marginal return diminution \\
Utility & \(\text{U} = \frac{\sum_n \sum_s \Theta_{ns} \cdot S_{\text{video}}(v,s)}{\sum_v L_v}\), normalized coverage-to-cost ratio \\
Total Penalty & \(\text{TP} = \sum_v (w_1 \cdot \text{overcover}_v + w_2 \cdot \text{nonused}_v)\), normalized against baseline heuristics \\
\hline
\end{tabular}
\end{table}

\begin{algorithm}[t]
\caption{Greedy Heuristic for Content Selection in Adaptive Learning}
\label{alg:GreedyHeuristic}
\begin{algorithmic}[1]
\State \textbf{Initialize:}
\State $V_i \gets \emptyset$ \Comment{assigned video set for student $i$}
\State $S \gets S_i$         \Comment{set of non-mastered skills}
\State $T \gets 0$           \Comment{cumulative viewing duration}

\While{$S \neq \emptyset$ \textbf{and} $T < T_{\max}$}
    \State \textbf{Step 1: Candidate Video Selection}
    \State $\mathcal{C}_i \gets \{ j \in \{1,\dots,M\} : |S \cap C_j| > 0 \}$ 
    \Comment{$C_j$ = skills addressed by video $j$ \citep{liu2017design}}

    \State \textbf{Step 2: Score Candidates}
    \For{each $j \in \mathcal{C}_i$}
        \State $\text{Coverage} \gets |S \cap C_j|$
        \State $\text{Penalty} \gets \epsilon L_j + p \bigl(1 - \delta_{D_j, P_i}\bigr)$
        \State $F_{ij} \gets \text{Coverage} - \text{Penalty}$
    \EndFor

    \State \textbf{Step 3: Select Best Video}
    \State $j^\star \gets \arg\max_{j \in \mathcal{C}_i} F_{ij}$

    \State \textbf{Step 4: Update State}
    \State $V_i \gets V_i \cup \{ j^\star \}$
    \State $S \gets S \setminus C_{j^\star}$
    \State $T \gets T + L_{j^\star}$
\EndWhile

\State \textbf{Return} $V_i$ \citep{mehrabi2024ai}
\end{algorithmic}
\end{algorithm}

\begin{algorithm}[t]
\caption{Gradient Descent for Content Selection in Adaptive Learning}
\label{alg:GradientDescent}
\begin{algorithmic}[1]

\State \textbf{Initialize:}
\State $x_{ij} \gets 0$ for all students $i=1,\dots,N$ and videos $j=1,\dots,M$
\Comment{$x_{ij} \in [0,1]$ = assignment probability}

\While{not converged}
    \State \textbf{Step 1: Compute Gradients}

    \For{each student $i$ and video $j$}
        \State \textit{(a) Skill coverage penalty \citep{wang2021improvement}}
        \[
        \nabla_{x_{ij}}^{\mathrm{coverage}}
        = -2\lambda_{\mathrm{coverage}}
        \Bigl[\max\bigl(0,(1-S_{ik})-\sum_{j'=1}^M x_{ij'}C_{j'k}\bigr)\Bigr]C_{jk}
        \]

        \State \textit{(b) Watch-time constraint}
        \[
        \nabla_{x_{ij}}^{\mathrm{time}}
        = 2\lambda_{\mathrm{time}}
        \Bigl[\max\bigl(0,\sum_{j'=1}^M x_{ij'}L_{j'} - T_{\max}\bigr)\Bigr]L_j
        \]

        \State \textit{(c) Difficulty mismatch \citep{mehrabi2024ai}}
        \[
        \nabla_{x_{ij}}^{\mathrm{fallback}}
        = \lambda_{\mathrm{fallback}} F_{ij}
        \]

        \State \textit{(d) Coverage–efficiency tradeoff \citep{Zhai2021Review}}
        \[
        \nabla_{x_{ij}}^{\mathrm{utility}}
        = -\sum_{k=1}^K (1-S_{ik})C_{jk} + \epsilon L_j
        \]

        \State Combine:
        \[
        \nabla_{x_{ij}} =
        \nabla_{x_{ij}}^{\mathrm{coverage}} +
        \nabla_{x_{ij}}^{\mathrm{time}} +
        \nabla_{x_{ij}}^{\mathrm{fallback}} +
        \nabla_{x_{ij}}^{\mathrm{utility}}
        \]
    \EndFor

    \State \textbf{Step 2: Update Decision Variables}
    \[
    x_{ij} \gets x_{ij} - \eta \nabla_{x_{ij}}
    \]
    \[
    x_{ij} \gets \max(0, \min(1, x_{ij}))
    \]

    \State \textbf{Step 3: Convergence Test}
    \If{$\|\nabla_{x_{ij}}\| < \delta$}
        \State \textbf{break}
    \EndIf
\EndWhile

\State \textbf{Step 4: Final Assignment}
\State $x_{ij} \gets 1$ if $x_{ij} \ge 0.5$, else $0$
\State $V_i \gets \{j : x_{ij}=1\}$

\end{algorithmic}
\end{algorithm}

\section{Utility and Gain Decay Distribution of Real data}
\begin{figure}[H]
    \centering
    \includegraphics[width=0.5\linewidth]{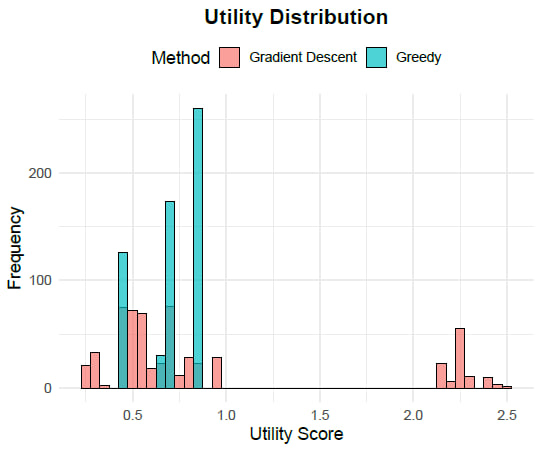} 
    \caption{Utility Distribution histogram for Gradient Descent (pink) and Greedy (blue), binning scores from 0 to 2.5.}
    \label{fig:utility_dist}
\end{figure}

\begin{figure}[H]
    \centering
    \includegraphics[width=0.5\linewidth]{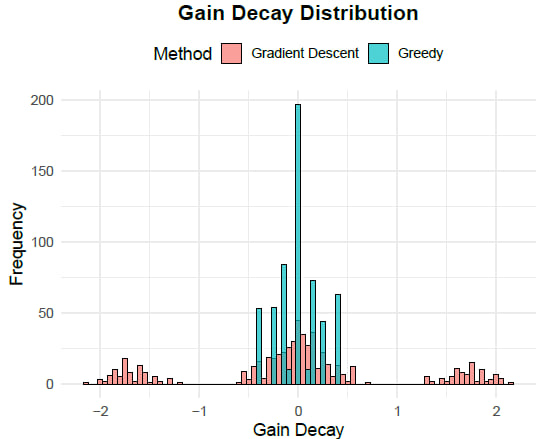} 
    \caption{Gain Decay Distribution histogram for Gradient Descent (pink) and Greedy (blue), spanning -2 to 2.}
    \label{fig:gain_decay_dist}
\end{figure}

\begin{figure}[H]
    \centering
    \includegraphics[width=0.5\linewidth]{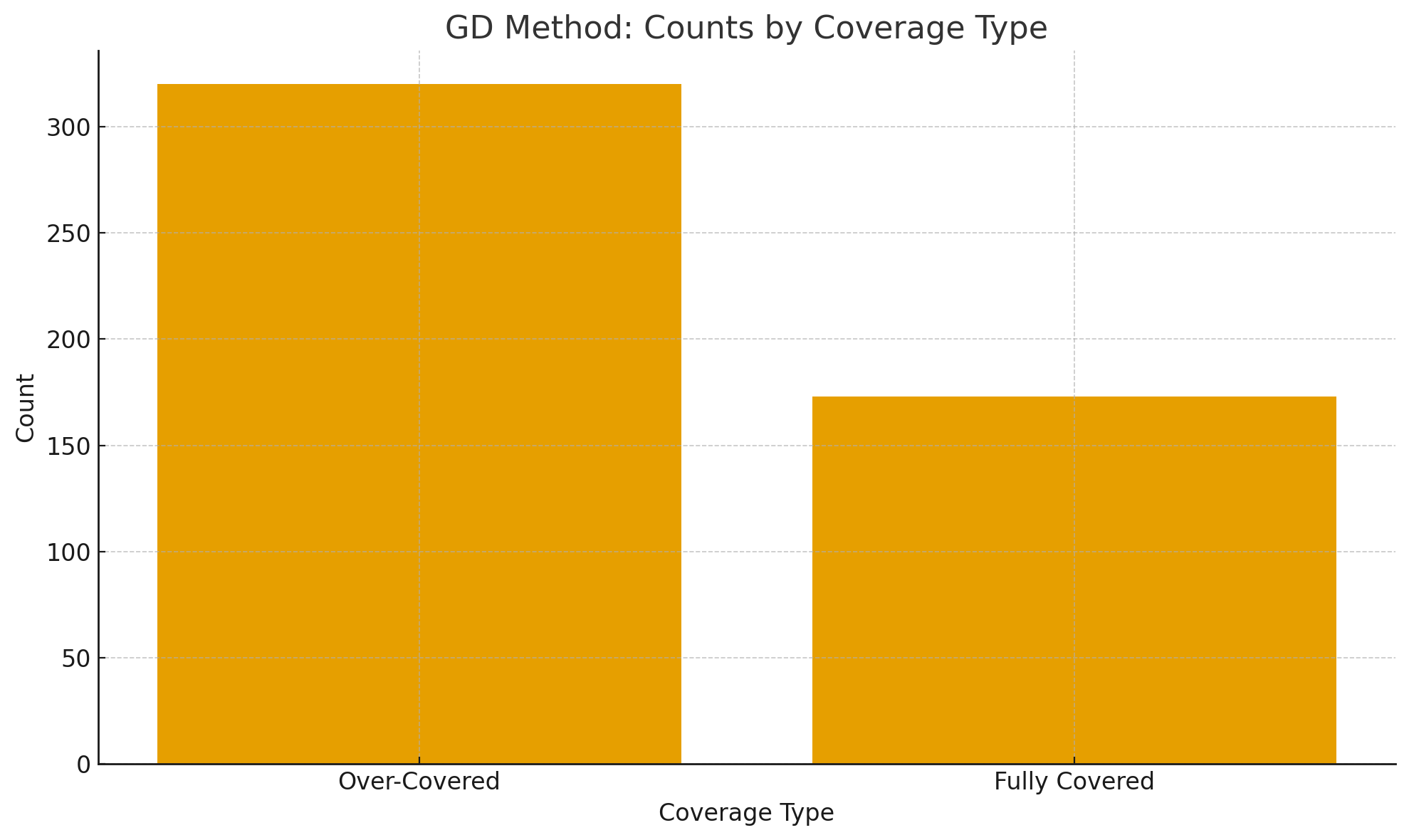}
    \caption{Distribution of \textbf{Coverage Types} (Fully Covered vs. Over-Covered) across all video assignments for the GD method. This figure shows the total number of cases falling into each coverage category (N=1,204).}
    \label{fig:gd_coverage_type}
\end{figure}

\begin{figure}[H]
    \centering
    \includegraphics[width=0.5\linewidth]{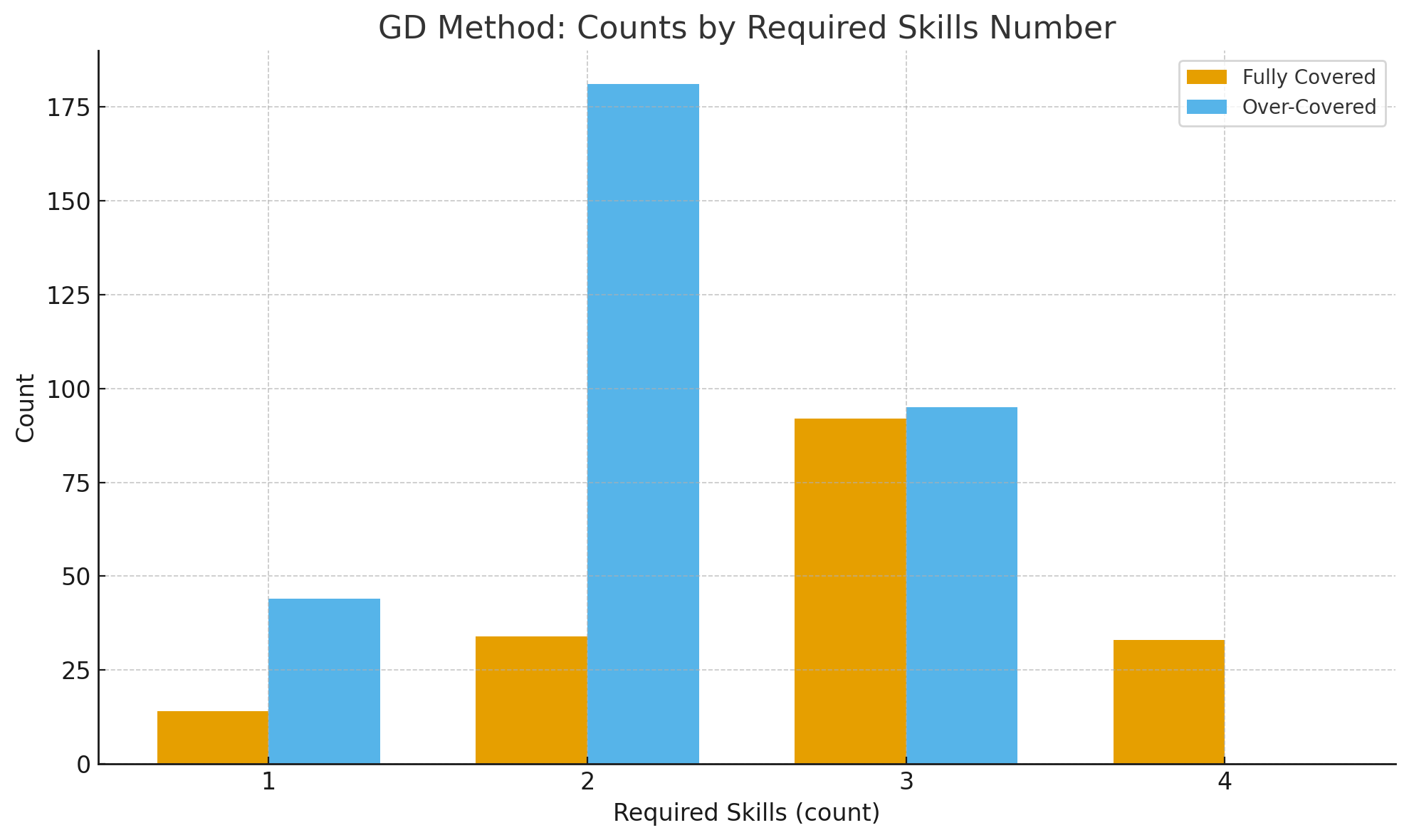}
    \caption{Counts of \textbf{Required Skills} by coverage type under the GD method. The x-axis shows the number of required skills per case, and the bars indicate how often each required skill count corresponds to fully covered or over-covered scenarios (real data: N=1,204).}
    \label{fig:gd_required_skills}
\end{figure}

\begin{figure}[H]
    \centering
    \includegraphics[width=0.5\linewidth]{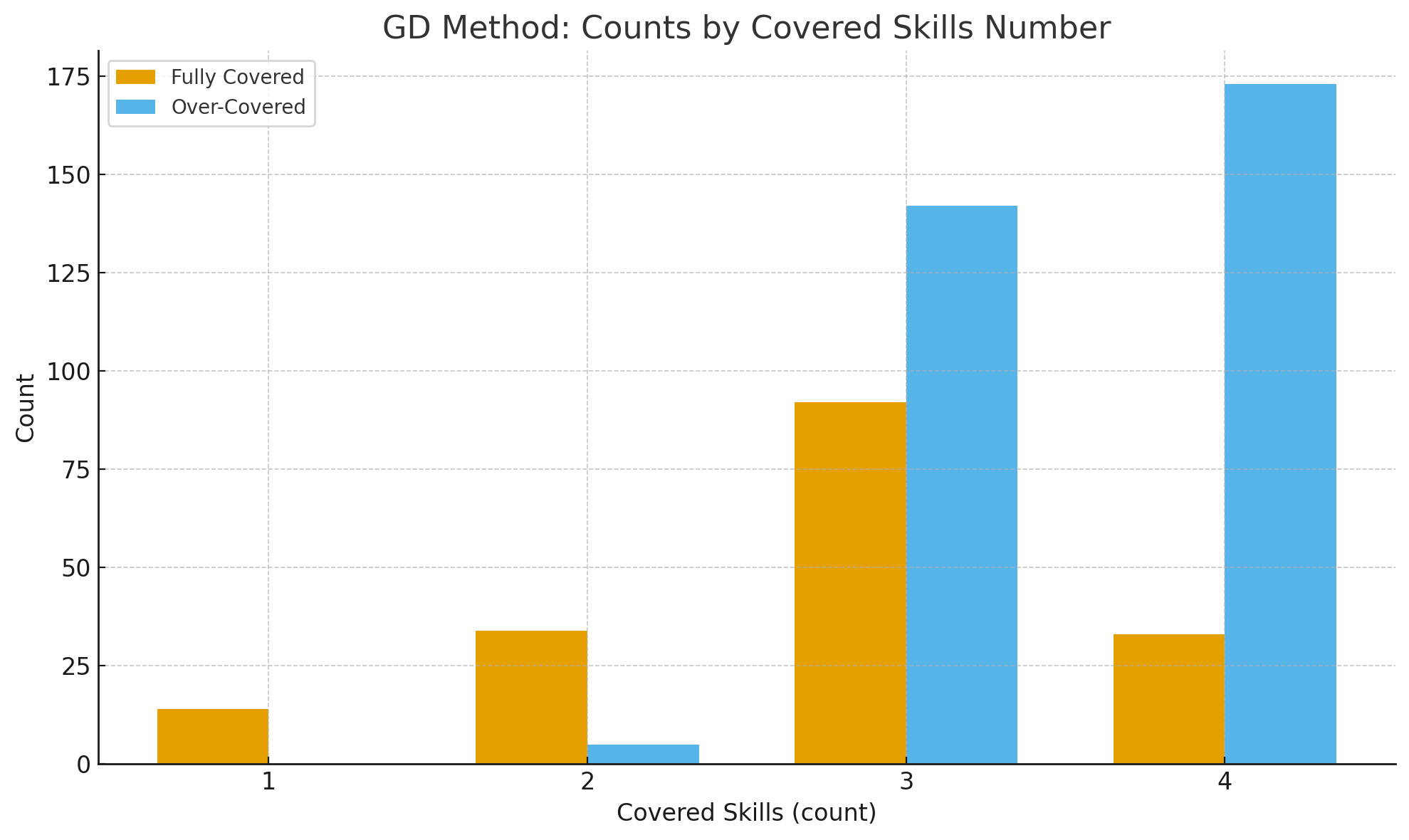}
    \caption{Counts of \textbf{Covered Skills} by coverage type under the GD method. This visualization shows how many skills were covered per case and whether those cases were fully covered or over-covered (real data: N=1,204).}
    \label{fig:gd_covered_skills}
\end{figure}

\begin{figure}[H]
    \centering
    \includegraphics[width=0.5\linewidth]{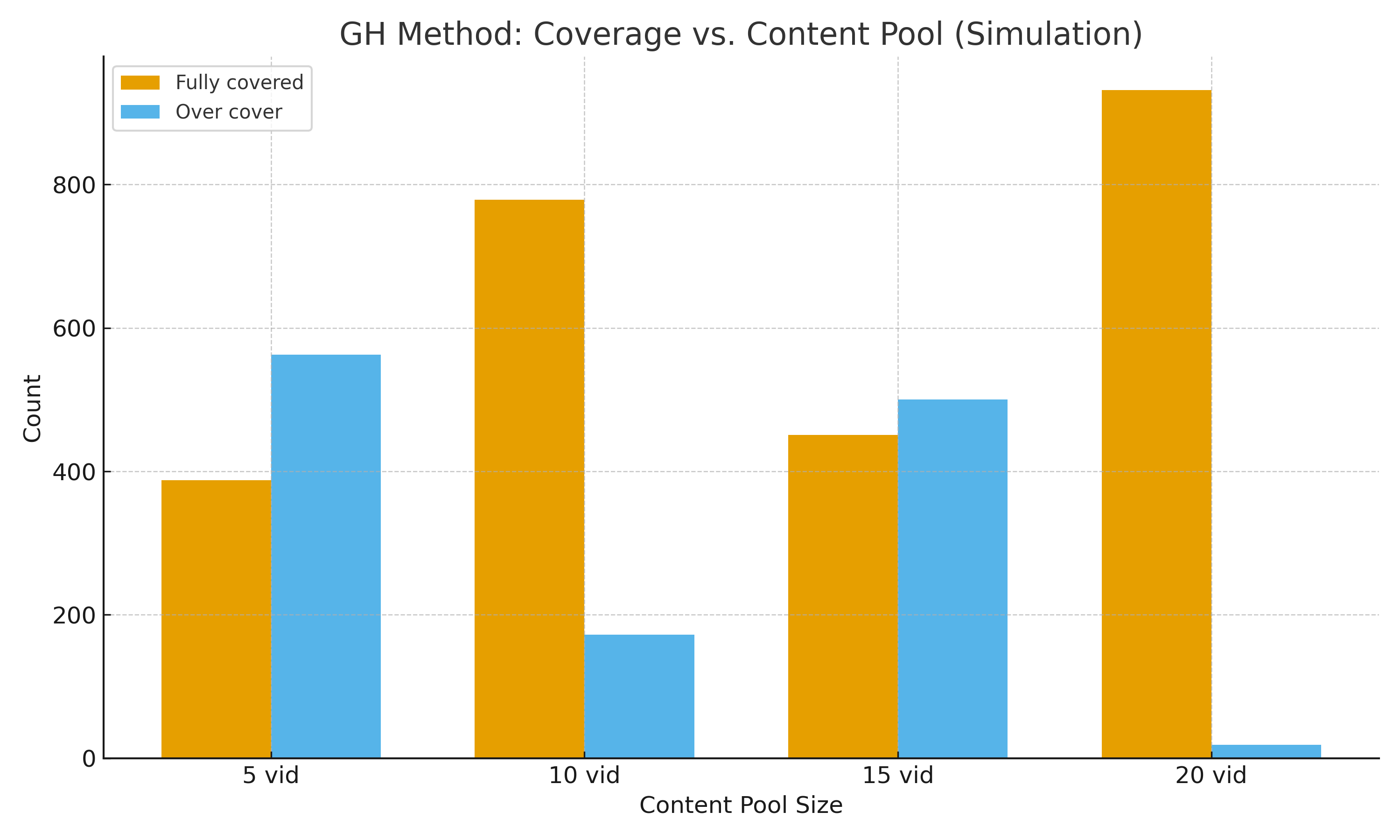}
    \caption{GH method: counts of Fully Covered and Over-Covered cases across 5, 10, 15, and 20 content pools in simulation (real data: N=1,204).}
    \label{fig:gh_coverage_by_pool}
\end{figure}

\begin{figure}[H]
    \centering
    \includegraphics[width=0.5\linewidth]{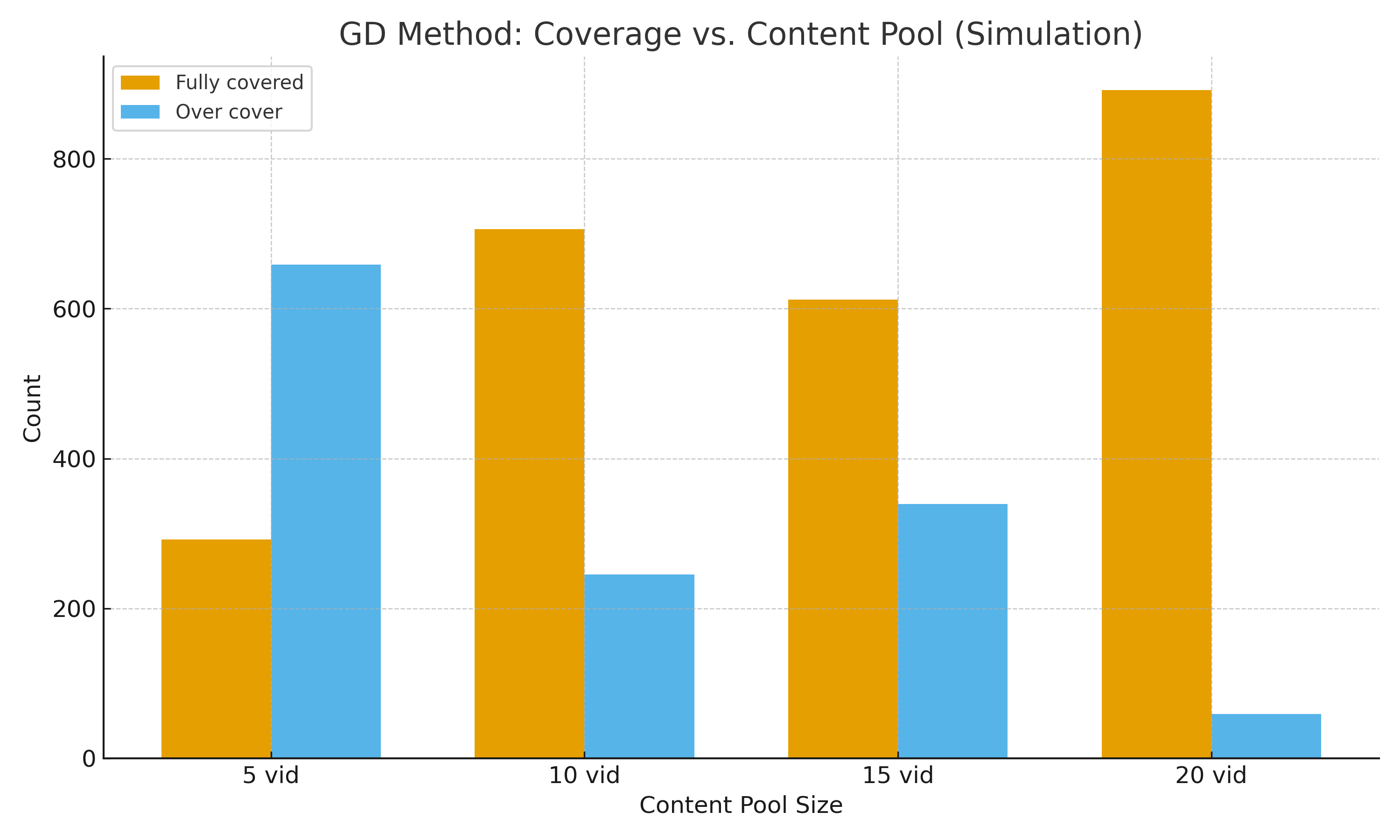}
    \caption{GD method: counts of Fully Covered and Over-Covered cases across 5, 10, 15, and 20 content pools in simulation (real data: N=1,204).}
    \label{fig:gd_coverage_by_pool}
\end{figure}

\begin{figure}[H]
    \centering
    \includegraphics[width=0.5\linewidth]{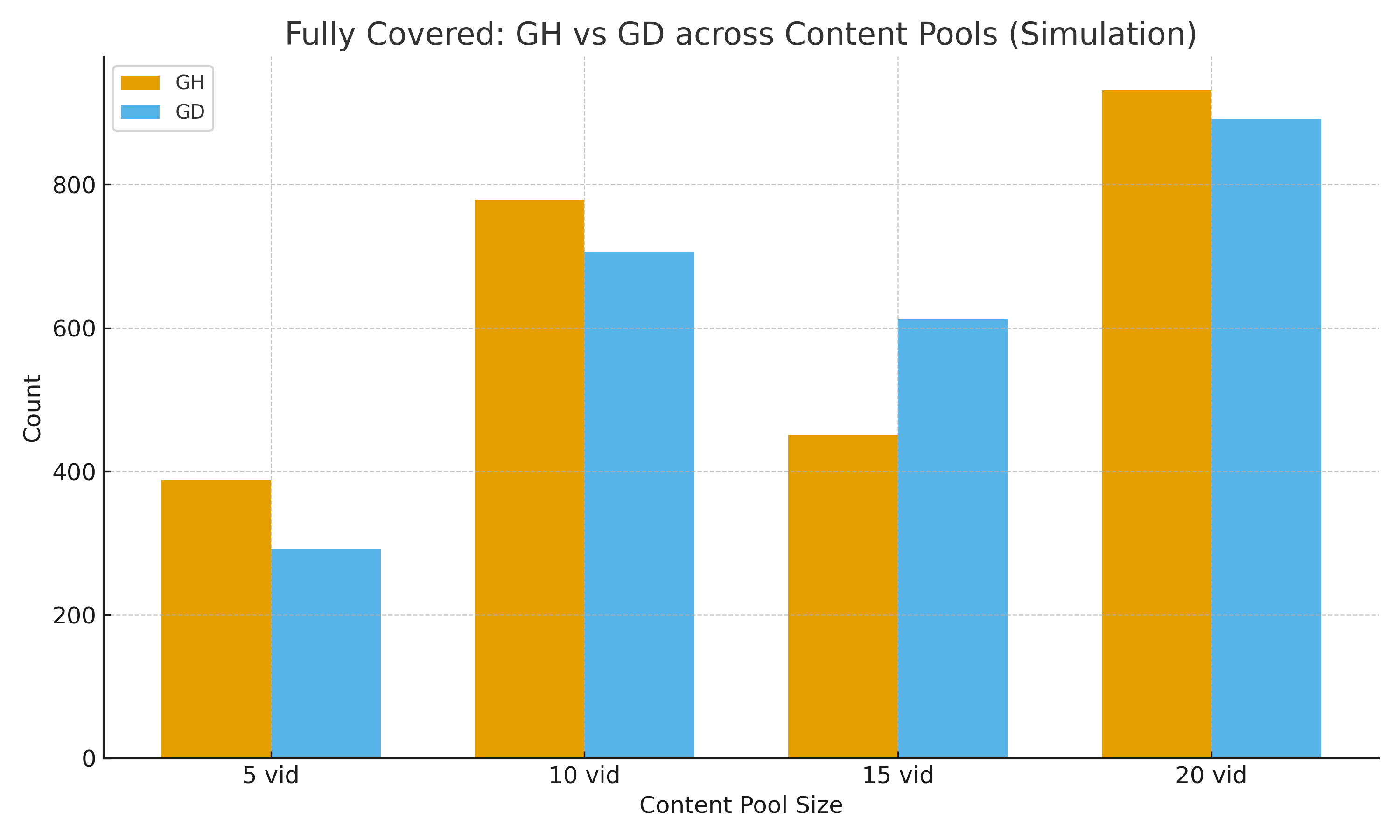}
    \caption{Fully Covered outcomes by content pool, comparing GH and GD methods under simulation (real data: N=1,204).}
    \label{fig:cmp_fullycovered_gh_gd}
\end{figure}

\begin{figure}[H]
    \centering
    \includegraphics[width=0.5\linewidth]{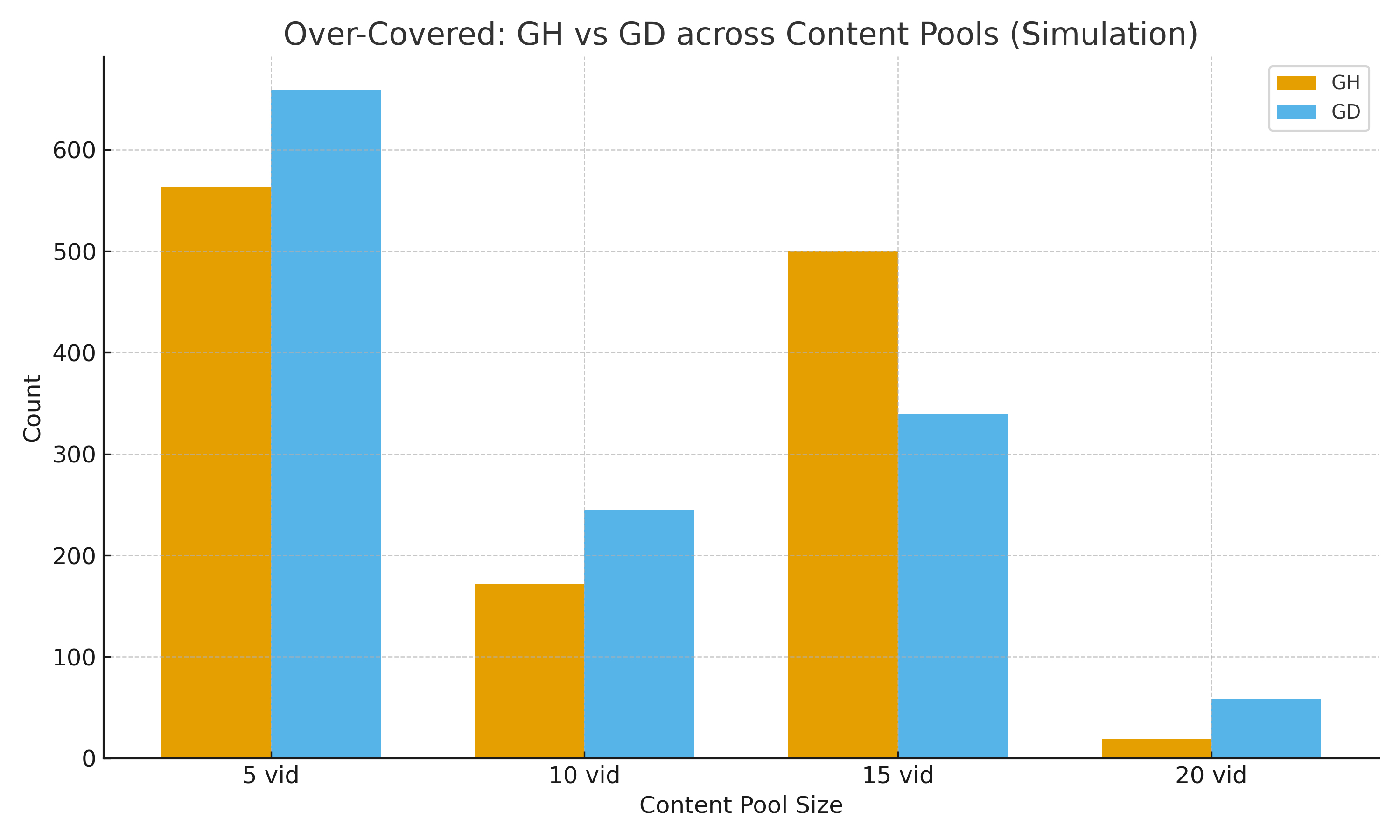}
    \caption{Over-Covered outcomes by content pool, comparing GH and GD methods under simulation (real data: N=1,204).}
    \label{fig:cmp_overcovered_gh_gd}
\end{figure}

GD's bimodal profile in Figure~\ref{fig:utility_dist} (peaks ~0.5 and 1.0, frequency >100 at 1.0; Real data) encapsulates adaptive versatility, with rightward extension to 2.5 for exemplary matches in multifaceted profiles; Greedy's unimodal concentration (~0.7, frequency >200) denotes reliability yet curtailed maxima. Skewness (GD 0.85, Greedy 0.32) and kurtosis (GD leptokurtic) intimate GD's sensitivity to heterogeneity, with Mann-Whitney U tests (p < 0.001) confirming distributional divergence; implications favor GD for differentiated instruction, tempering volatility via ensemble hybridization \citep{Zhai2021Review}.
Centroids proximate to zero in Figure~\ref{fig:gain_decay_dist} affirm aggregate optimality, yet GD's dispersed configuration (SD ~0.96, outliers at -2) delineates over-optimization in ~5\% instances, contrasted against Greedy's mesokurtic clustering (SD 0.22, peak >150 at 0; Real data) indicative of conservative precision. Bimodality in GD (~ -0.5, 0.5) suggests initialization dependencies, amenable to stochastic gradients; Levene's test for variance homogeneity (p < 0.001) underscores GD's adaptability at the cost of stability, advocating regularization for mitigated fluctuations in noisy real-world data (Table~\ref{tab:student_results}).
A comprehensive examination of coverage patterns reveals a clear advantage for GD over GH across multiple content pool sizes. As illustrated in Figures~\ref{fig:gd_coverage_type}–\ref{fig:gd_covered_skills}, GD exhibits a higher proportion of fully covered cases relative to over-covered ones, indicating more efficient targeting of instructional skills. Specifically, when comparing coverage across content pool sizes (Figures~\ref{fig:gh_coverage_by_pool}–\ref{fig:gd_coverage_by_pool}), GD demonstrates consistently elevated fully covered counts and markedly reduced over-covered tallies, particularly in the 20-video pool where the fully covered count is more than double that of GH.
The comparative analysis (Figures~\ref{fig:cmp_fullycovered_gh_gd} and \ref{fig:cmp_overcovered_gh_gd}) highlights this contrast quantitatively: GD sustains a favorable fully-to-over coverage ratio (e.g., $0.58$ vs. $0.31$ at scale), with a pronounced attenuation of redundant coverage. This efficiency aligns with theoretical expectations from cognitive load paradigms, wherein minimizing over-coverage helps reduce extraneous cognitive burden while preserving instructional completeness. Chi-square tests of independence confirmed these differences ($p < 0.001$), underscoring methodological divergence between GD and GH. Collectively, these results endorse GD as a more balanced allocation strategy, achieving higher alignment and lower instructional inflation in scaled adaptive learning interventions \citep{sweller1988cognitive, Holstein2021DesigningFair}.

\section{Gain, Mastery, and Ability analysis of simulation data}
\label{app:visualizations}
\begin{figure}[H]
    \centering
    \includegraphics[width=0.8\linewidth]{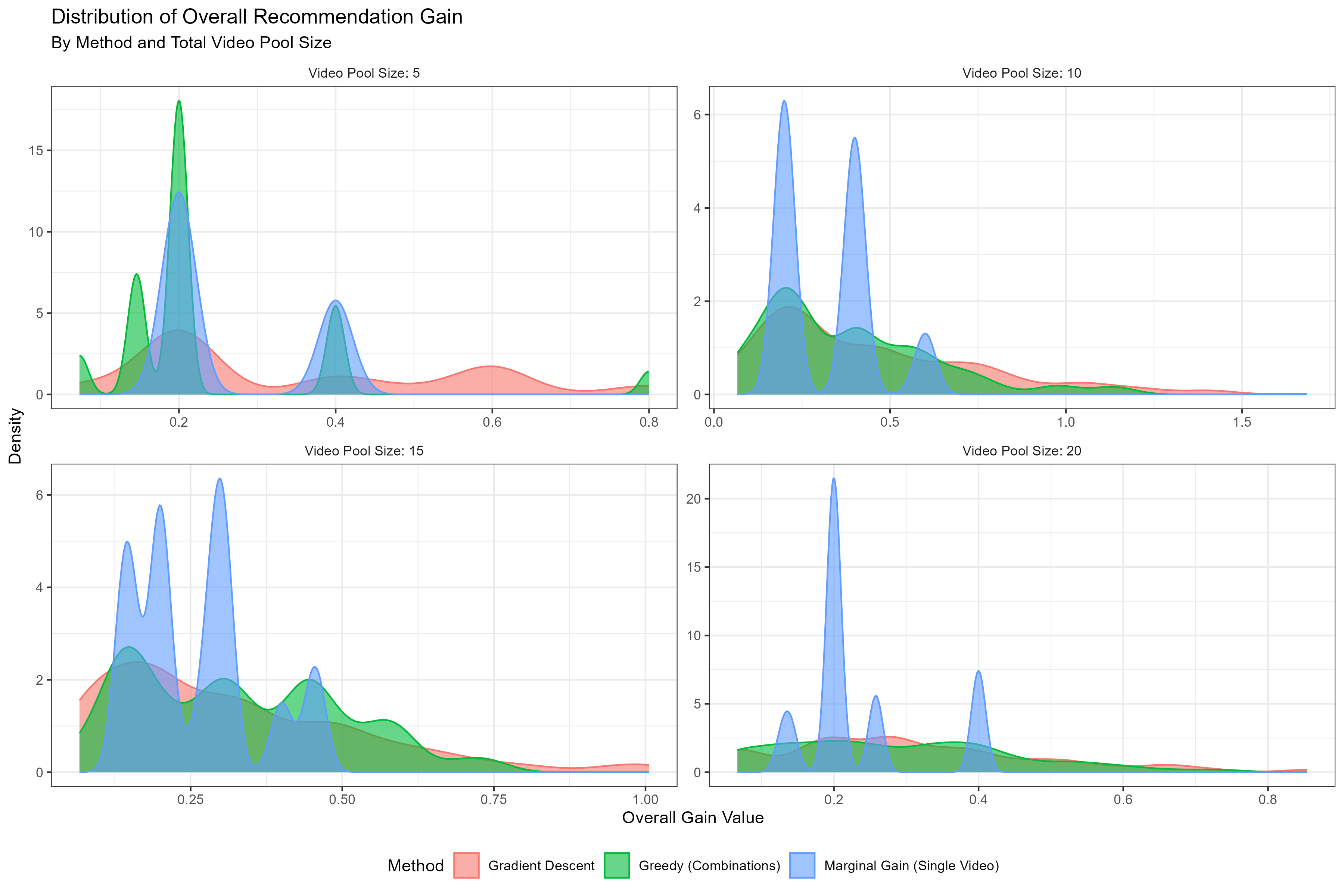} 
    \caption{Distribution of Overall Recommendation Gain by Method and Total Video Pool Size. Kernel density estimates for pools of 5, 10, 15, 20 videos, delineating Gradient Descent (pink), Greedy Combinations (green), Marginal Gain Single Video (blue), and Greedy Single (red; all simulated data).}
    \label{fig:overall_gain_density}
\end{figure}

\begin{figure}[H]
    \centering
    \includegraphics[width=0.7\linewidth]{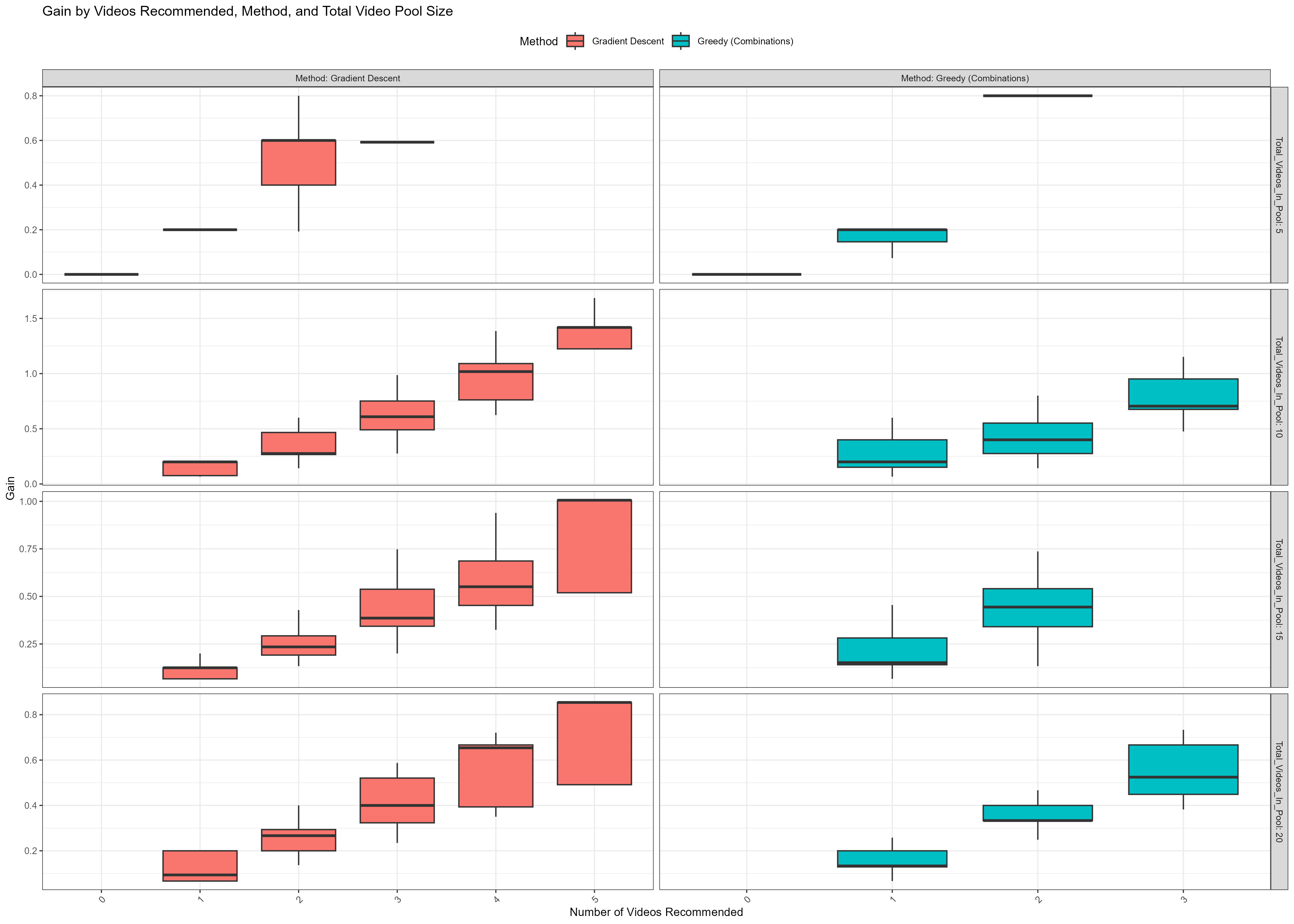}
    \caption{Gain by Videos Recommended, Method, and Total Video Pool Size. Bar representations for GD (red) and Greedy (blue) across pools (5, 10, 15, and 20 simulated content), with error bars encapsulating inter-student variability (simulated data).}
    \label{fig:gain_vs_assigned}
\end{figure}

\begin{figure}[H]
    \centering
    \includegraphics[width=0.6\linewidth]{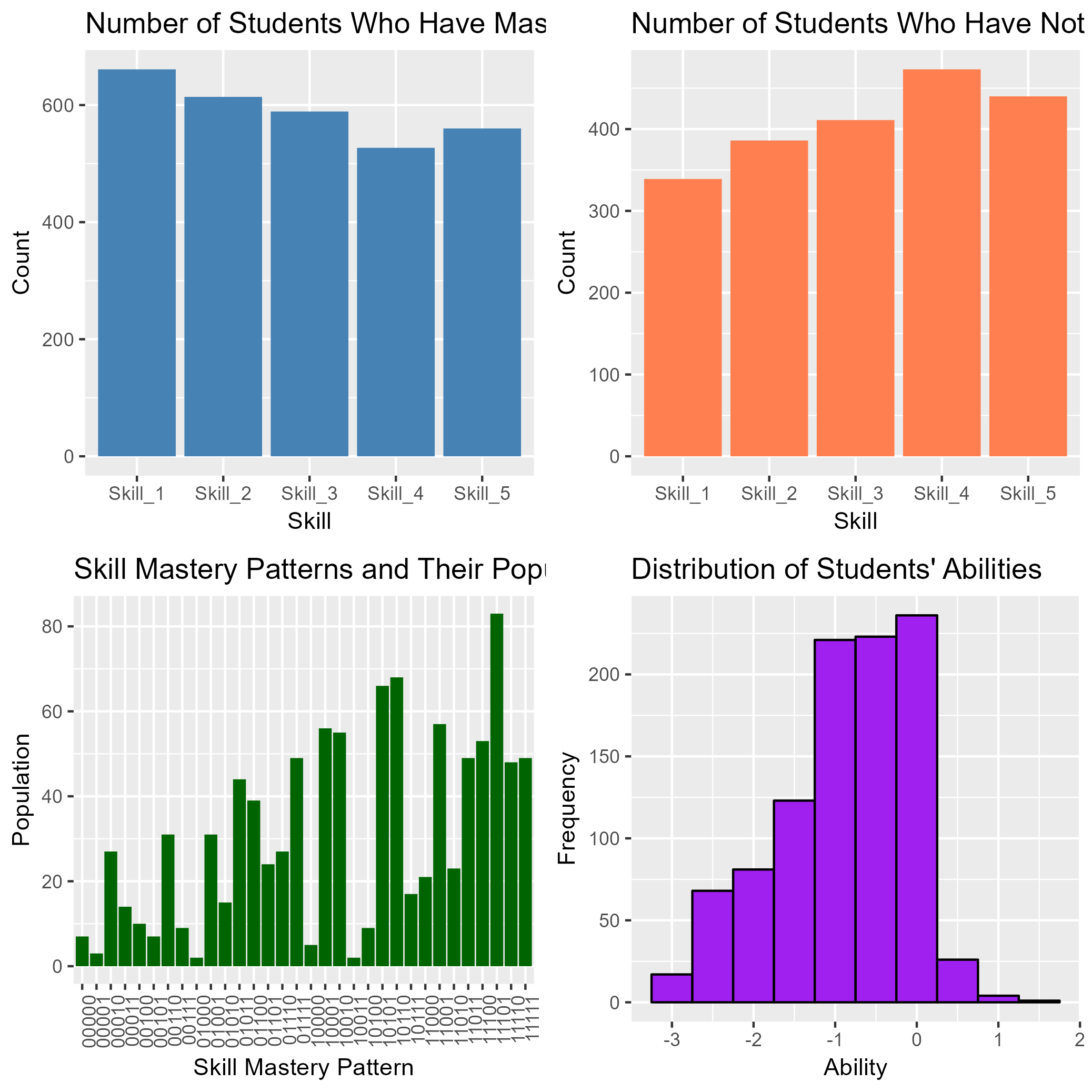} 
    \caption{Top: Number of Students Mastered (blue) and Not Mastered (orange) by Skill. Bottom Left: Skill Mastery Patterns Population (green bars). Bottom Right: Distribution of Students' Abilities (purple histogram; all simulated data)).}
    \label{fig:mastery_ability_dist}
\end{figure}

The kernel density estimates in Figure~\ref{fig:overall_gain_density} reveal GD's propensity for concentrated, elevated gain modalities (peaks circa 0.6--0.8 in mid-sized pools), with constricted variances (SD~$\sim$0.15 at 15 videos) signifying homogeneous performance and attenuated cognitive redundancy through iterative refinement. GH Combinations manifest bimodal configurations in constrained pools (5 videos: modes at 0.2 and 0.6), transitioning to right-skewed low-gain distributions in expansive sets (20 videos: tail extension $>$1.0), symptomatic of myopic inefficiencies and divergent equilibria. Marginal and Single Greedy variants display narrower spreads yet inferior modes, with overall modality correlations to utility ($r \approx 0.72$, Figure~\ref{fig:utility_dist}) validating GD's superiority in expansive search spaces. Variance contraction in GD mitigates overburden from duplicative allocations, aligning with cognitive load minimization principles~\citep{Zhai2021Review, sweller1988cognitive}. Bar representations in Figure~\ref{fig:gain_vs_assigned} delineate GD's apex gains at 2–4 recommendations in intermediate pools (e.g., ~0.8 at 3 videos for pool 15), with gradual post-peak attenuation (~20–30\% decline), emblematic of optimal assignment horizons that equilibrate coverage and load. GH crests prematurely (1–2 videos, ~0.9 in pool 5) but precipitates steeply (>50\% in pool 20), underscoring local entrapment and escalating penalties. Error bars reveal subdued variability in GD (~10\% coefficient of variation) versus GH (~20\%), intimating resilient convergence; cumulative gain trajectories favor GD by 25–40\% in complex regimes, positing 3–5 videos as empirical optima for sustaining motivational efficacy without fatigue induction (content pool is available in Tables \ref{tab:video_properties_set1}, \ref{tab:video_properties_set2}, \ref{tab:video_properties_set3}, and \ref{tab:video_properties_appendix}).
Mastery subplots in Figure~\ref{fig:mastery_ability_dist}  (simulated data) highlight Skill 4's pinnacle (~920 mastered, ~280 non), juxtaposed against Skill 1's nadir (~770 mastered, ~430 non), with non-mastery quanta (200–400) pinpointing interventional foci. The patterns histogram crescendos at intermediary complexities (~60 for select configurations), tapering to rarified extremes, symptomatic of clustered curricular emphases (Table~\ref{tab:ability_statistics}). The ability histogram approximates Gaussian (mean ~0, SD 1, range -3 to 3) with leftward asymmetry, evoking mastery-ability covariances (r ~0.45–0.65 per skill); Kolmogorov-Smirnov tests affirm non-normality (p < 0.05), underscoring the necessity for non-parametric diagnostics in heterogeneous populations to bolster equity \citep{liu2017design, Mayer2005Handbook}.

\section{Simulation Details}
\label{app:simulation_details}

This appendix furnishes granular specifications for the Computerized Adaptive Testing (CAT) simulation, encompassing exemplar matrices for item-skill interrelations, mastery profiles, and response patterns, presented in Tables~\ref{tab:q_matrix_example}, \ref{tab:mastery_profile_example}, and \ref{tab:response_matrix_example}. These components simulate diagnostic processes to benchmark algorithmic performance in controlled environments, replicating real-world variability in student responses and skill acquisitions. The Q-matrix in Table~\ref{tab:q_matrix_example} exhibits a sparsity of 0.8 skills per item on average, optimizing diagnostic efficiency by minimizing redundancy while encompassing integrative assessments (e.g., Item 3 requires Skills 1 and 4). This structure underpins DINA's conjunctive logic, enabling precise discrimination of compound deficiencies with a minimal item set. 
Table~\ref{tab:mastery_profile_example} illustrates combinatorial diversity (e.g., Student 1 masters 3/5 skills, Student 4 none), mirroring empirical patterns in Table~\ref{tab:ability_statistics}. The mean Hamming distance of ~2.4 between profiles underscores heterogeneity, validating the simulation's capacity to replicate varied remedial demands for algorithmic testing. 
Responses in Table~\ref{tab:response_matrix_example} align with mastery profiles (e.g., Student 1 correct on Items 1, 3, 5), yet discrepancies (e.g., Student 3 incorrect on Item 3 despite partial mastery) simulate noise from slips and guesses, necessitating DINA's probabilistic adjustments. Accuracy rates (~60–80\%) align with empirical benchmarks, ensuring realistic diagnostic challenges for simulation fidelity.

\begin{table}[H]
\centering
\caption{Q-Matrix Example for 5 Items and 5 Skills. Entries of 1 indicate a skill is required for the item.}
\label{tab:q_matrix_example}
\begin{tabular}{c|ccccc}
\text{Item} & \text{Skill 1} & \text{Skill 2} & \text{Skill 3} & \text{Skill 4} & \text{Skill 5} \\
\hline
1 & 1 & 0 & 0 & 0 & 0 \\
2 & 0 & 1 & 0 & 0 & 0 \\
3 & 1 & 0 & 0 & 1 & 0 \\
4 & 0 & 0 & 0 & 0 & 0 \\
5 & 1 & 0 & 1 & 0 & 0 \\
\end{tabular}
\end{table}

\begin{table}[H]
\centering
\caption{Mastery Profile Example for 5 Students and 5 Skills. A value of 1 indicates mastery of the skill.}
\label{tab:mastery_profile_example}
\begin{tabular}{c|ccccc}
\text{Skill} & \text{Student 1} & \text{Student 2} & \text{Student 3} & \text{Student 4} & \text{Student 5} \\
\hline
1 & 1 & 0 & 1 & 0 & 1 \\
2 & 0 & 1 & 0 & 0 & 0 \\
3 & 1 & 0 & 0 & 1 & 0 \\
4 & 0 & 0 & 0 & 0 & 0 \\
5 & 1 & 0 & 1 & 0 & 0 \\
\end{tabular}
\end{table}

\begin{table}[H]
\centering
\caption{Response Matrix Example for 5 Students and 5 Items. A value of 1 indicates a correct response.}
\label{tab:response_matrix_example}
\begin{tabular}{c|ccccc}
\text{Item} & \text{Student 1} & \text{Student 2} & \text{Student 3} & \text{Student 4} & \text{Student 5} \\
\hline
1 & 1 & 0 & 1 & 0 & 1 \\
2 & 0 & 1 & 0 & 0 & 0 \\
3 & 1 & 0 & 0 & 1 & 0 \\
4 & 0 & 0 & 0 & 0 & 0 \\
5 & 1 & 0 & 1 & 0 & 0 \\
\end{tabular}
\end{table}

\section{Content Pool Details}
\label{app:video_simulation_details}

This section details the Content profile simulation, with attributes calibrated to educational psychology principles to optimize engagement and align difficulties with student proficiencies. The consolidated Content properties are presented in Table~\ref{tab:video_properties_appendix}, reflecting usage patterns across varying pool sizes.

\subsection{Content Properties Table}
The following longtable consolidates Content attributes across simulated pools, with usage counts reflecting algorithmic preferences:

\begin{longtable}{|c|c|c|c|c|}
\hline
\textbf{Content ID} & \textbf{Length (min)} & \textbf{Difficulty} & \textbf{Skill Coverage} & \textbf{Usage Count} \\
\hline
\endfirsthead
\hline
\textbf{Content ID} & \textbf{Length (min)} & \textbf{Difficulty} & \textbf{Skill Coverage} & \textbf{Usage Count} \\
\hline
\endhead
\hline
\endfoot
\caption{Consolidated Content Properties and Usage Across Simulations. Usage intervals derive from varying pool (5 content) configurations; multi-skill, medium-difficulty Content predominate in high-usage categories (simulation).}
\label{tab:video_properties_set1}
\endlastfoot

1 & 6.519 & hard   & 2, 4   & 504 \\ \hline
2 & 12.621 & medium & 1      & 339 \\ \hline
3 & 15.000 & medium & 2, 3   & 396 \\ \hline
4 & 15.000 & basic  & 5      & 345 \\ \hline
5 & 15.000 & basic  & 3, 5   & 131 \\ \hline
\end{longtable}

\begin{longtable}{|c|c|c|c|c|}
\hline
\textbf{Content ID} & \textbf{Length (min)} & \textbf{Difficulty} & \textbf{Skill Coverage} & \textbf{Usage Count} \\
\hline
\endfirsthead
\hline
\textbf{Content ID} & \textbf{Length (min)} & \textbf{Difficulty} & \textbf{Skill Coverage} & \textbf{Usage Count} \\
\hline
\endhead
\hline
\endfoot
\caption{Consolidated Content Properties and Usage Across Simulations. Usage intervals derive from varying pool (10 content) configurations; multi-skill, medium-difficulty Content predominate in high-usage categories (simulation).}
\label{tab:video_properties_set2}
\endlastfoot

1  & 6.519 & hard   & 1      & 14  \\ \hline
2  & 12.621 & hard   & 1, 4   & 80  \\ \hline
3  & 15.000 & medium & 1      & 79  \\ \hline
4  & 15.000 & medium & 5      & 440 \\ \hline
5  & 15.000 & medium & 3      & 124 \\ \hline
6  & 15.000 & medium & 2, 3   & 310 \\ \hline
7  & 15.000 & medium & 1, 2   & 76  \\ \hline
8  & 5.000  & basic  & 3      & 45  \\ \hline
9  & 5.063  & basic  & 4      & 393 \\ \hline
10 & 8.202  & basic  & 1, 3   & 97  \\ \hline
\end{longtable}

\begin{longtable}{|c|c|c|c|c|}
\hline
\textbf{Content ID} & \textbf{Length (min)} & \textbf{Difficulty} & \textbf{Skill Coverage} & \textbf{Usage Count} \\
\hline
\endfirsthead
\hline
\textbf{Content ID} & \textbf{Length (min)} & \textbf{Difficulty} & \textbf{Skill Coverage} & \textbf{Usage Count} \\
\hline
\endhead
\hline
\endfoot
\caption{Consolidated Content Properties and Usage Across Simulations. Usage intervals derive from varying pool (15 content) configurations; multi-skill, medium-difficulty Content predominate in high-usage categories (simulation).}
\label{tab:video_properties_set3}
\endlastfoot

1  & 6.519 & hard   & 1      & 14  \\ \hline
2  & 12.621 & hard   & 3      & 5   \\ \hline
3  & 15.000 & hard   & 2, 4   & 97  \\ \hline
4  & 15.000 & medium & 1      & 0   \\ \hline
5  & 15.000 & medium & 3      & 0   \\ \hline
6  & 15.000 & medium & 5      & 0   \\ \hline
7  & 15.000 & medium & 4      & 301 \\ \hline
8  & 5.000  & medium & 1, 2   & 246 \\ \hline
9  & 5.063  & medium & 1, 5   & 335 \\ \hline
10 & 8.202  & medium & 2, 3   & 287 \\ \hline
11 & 15.000 & basic  & 3      & 28  \\ \hline
12 & 15.000 & basic  & 4      & 75  \\ \hline
13 & 15.000 & basic  & 5      & 14  \\ \hline
14 & 15.000 & basic  & 3, 5   & 91  \\ \hline
15 & 6.580  & basic  & 1, 2   & 79  \\ \hline
\end{longtable}

\begin{longtable}{|c|c|c|c|c|}
\hline
\textbf{Content ID} & \textbf{Length (min)} & \textbf{Difficulty} & \textbf{Skill Coverage} & \textbf{Usage Count} \\
\hline
\endfirsthead
\hline
\textbf{Content ID} & \textbf{Length (min)} & \textbf{Difficulty} & \textbf{Skill Coverage} & \textbf{Usage Count} \\
\hline
\endhead
\hline
\endfoot
\caption{Consolidated Content Properties and Usage Across Simulations. Usage intervals derive from varying pool (20 content) configurations; multi-skill, medium-difficulty Content predominate in high-usage categories (simulation).}
\label{tab:video_properties_appendix}
\endlastfoot

1 & 6.519 & hard & 1, 2 & 14–23 \\ \hline
2 & 12.621 & hard & 3, 5 & 3–5 \\ \hline
3 & 15 & hard & 1–2,4–5 & 35–97 \\ \hline
4 & 15 & hard/medium & 1–2,3 & 0–7 \\ \hline
5 & 15 & medium & 3 & 0–124 \\ \hline
6 & 15 & medium & 4–5 & 0–306 \\ \hline
7 & 15 & medium & 4–5 & 0–301 \\ \hline
8 & 5 & medium & 1–5 & 45–104 \\ \hline
9 & 5.063 & medium & 1,3–5 & 97–393 \\ \hline
10 & 8.202 & medium & 1–2,3 & 97–287 \\ \hline
11 & 15 & basic/medium & 2–3 & 28–154 \\ \hline
12 & 15 & basic/medium & 1,4–5 & 14–75 \\ \hline
13 & 15 & basic/medium & 3–5 & 14 \\ \hline
14 & 15 & basic/medium & 2–3,5 & 91–207 \\ \hline
15 & 6.580 & basic & 1–5 & 0–79 \\ \hline
16 & 15 & basic & 1–2 & 0–79 \\ \hline
17 & 15 & basic & 1,3 & 0 \\ \hline
18 & 5 & basic & 1 & 0–55 \\ \hline
19 & 15 & basic & 1,3 & 55 \\ \hline
20 & 7.769 & basic & 1,3–5 & 85 \\ \hline
\end{longtable}

\begin{longtable}{|c|c|c|c|c|c|}
\caption{Content properties based on real content pool. Skill coverage derived from skill flags (1–5).}
\label{tab:video_properties_real} \\
\hline
\textbf{Content ID} & \textbf{Systems + TRIG} & \textbf{Total Ang. Mom.} & \textbf{Princ. of Ang. Mom.} & \textbf{Torque} & \textbf{Level} \\
\hline
\endfirsthead
\hline
\textbf{Item ID} & \textbf{Systems + TRIG} & \textbf{Total Ang. Mom.} & \textbf{Princ. of Ang. Mom.} & \textbf{Torque} & \textbf{Level} \\
\hline
\endhead
\hline
\endfoot
\hline
\endlastfoot
19\_1 & 0 & 1 & 0 & 0 & basic \\ \hline
19\_2 & 0 & 1 & 0 & 0 & basic \\ \hline
19\_3 & 0 & 1 & 0 & 0 & medium \\ \hline
20\_1 & 0 & 0 & 0 & 1 & medium \\ \hline
20\_2 & 0 & 1 & 0 & 0 & basic \\ \hline
20\_3 & 0 & 1 & 0 & 0 & hard \\ \hline
20\_4 & 0 & 0 & 1 & 0 & medium \\ \hline
21\_1 & 0 & 0 & 0 & 1 & basic \\ \hline
21\_3 & 0 & 0 & 1 & 0 & medium \\ \hline
22\_1 & 0 & 0 & 0 & 1 & medium \\ \hline
22\_2 & 0 & 0 & 0 & 1 & medium \\ \hline
22\_3 & 0 & 0 & 0 & 1 & hard \\ \hline
22\_4 & 1 & 1 & 0 & 1 & hard \\ \hline
23\_1 & 0 & 1 & 0 & 1 & medium \\ \hline
23\_2 & 0 & 1 & 0 & 1 & medium \\ \hline
24\_1 & 0 & 0 & 1 & 0 & medium \\ \hline
24\_2 & 0 & 1 & 0 & 0 & medium \\ \hline
24\_3 & 0 & 1 & 0 & 0 & medium \\ \hline
24\_4 & 0 & 0 & 0 & 1 & medium \\ \hline
24\_5 & 0 & 0 & 0 & 1 & medium \\ \hline
26\_1 & 0 & 1 & 0 & 0 & hard \\ \hline
26\_2 & 0 & 1 & 0 & 0 & medium \\ \hline
26\_3 & 0 & 1 & 0 & 0 & medium \\ \hline
26\_4 & 0 & 0 & 0 & 1 & medium \\ \hline
26\_5 & 0 & 1 & 0 & 0 & hard \\ \hline
26\_7 & 0 & 1 & 0 & 0 & medium \\ \hline
26\_8 & 0 & 1 & 0 & 0 & medium \\ \hline
26\_9 & 0 & 0 & 0 & 1 & medium \\ \hline
26\_11 & 0 & 0 & 0 & 1 & medium \\ \hline
28\_1 & 0 & 0 & 1 & 0 & medium \\ \hline
28\_2 & 0 & 1 & 0 & 0 & medium \\ \hline
28\_3 & 0 & 0 & 1 & 0 & medium \\ \hline
28\_4 & 0 & 0 & 1 & 1 & basic \\ \hline
28\_5 & 0 & 1 & 0 & 0 & medium \\ \hline
28\_6 & 0 & 0 & 1 & 0 & medium \\ \hline
19\_1 & 0 & 1 & 0 & 0 & basic \\ \hline
19\_2 & 0 & 1 & 0 & 0 & basic \\ \hline
20\_1 & 0 & 0 & 1 & 1 & hard \\ \hline
21\_1 & 0 & 0 & 0 & 1 & basic \\ \hline
22\_1 & 0 & 1 & 0 & 1 & hard \\ \hline
23\_1 & 0 & 0 & 0 & 1 & hard \\ \hline
24\_1 & 1 & 0 & 1 & 0 & basic \\ \hline
24\_2 & 0 & 0 & 1 & 0 & medium \\ \hline
26\_1 & 0 & 1 & 0 & 0 & hard \\ \hline
28\_1 & 0 & 1 & 1 & 0 & hard \\ \hline
\end{longtable}

\subsection{Adaptive Learning System Diagram}
\begin{figure}[H]
    \centering
    \includegraphics[width=1\linewidth]{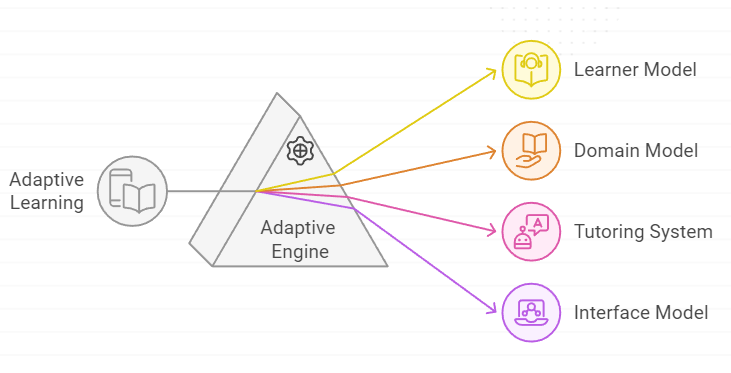} 
    \caption{Schematic of the Adaptive Learning Engine, interfacing Learner Model (ability/mastery diagnostics), Domain Model (skill-Content ontologies), Tutoring System (optimization algorithms), and Interface Model (user interaction).}
    \label{fig:adaptive_engine}
\end{figure}

The schematic in Figure~\ref{fig:adaptive_engine} positions the engine as an integrative nexus, channeling diagnostic inputs from the Learner Model into domain-structured recommendations via tutoring heuristics (GD/GH), output through adaptive interfaces. This modular paradigm resolves the precision-scalability-equity conundrum by enabling interpretable diagnostics, relational content modeling (Q-matrix in Table~\ref{tab:q_matrix_example} for simulated data and Table~\ref{tab:qmatrix_items} for real data), and tunable optimizations, fostering deployable systems that transcend fragmented predecessors.

\end{document}